\documentclass[a4paper]{amsart} 
\usepackage{amsmath,amsfonts,amssymb,graphicx,url,amsthm,hyperref}
\usepackage[left=1.9cm,right=1.9cm,top=3cm,bottom=3cm]{geometry}
\usepackage[nocompress]{cite}
\usepackage{xcolor}
\usepackage[noend]{algpseudocode}
\usepackage[ruled,vlined,norelsize,algosection]{algorithm2e}
\usepackage[T1]{fontenc}

\newcommand{\kernel}{\mathcal{K}}

\newcommand{\bs}{\boldsymbol}
\newcommand{\bm}{\bs}

\DeclareMathOperator{\spn}{span}

\usepackage{tikz,pgfplots,pgfplotstable}
\pgfplotsset{compat=newest}
\usetikzlibrary{arrows,snakes,backgrounds,positioning,calc,matrix}
\usepackage{graphicx,color,nicefrac} 
\usepackage{amssymb, amsmath, amsbsy,mathtools}
\usepackage{amsmath,amsfonts}
\usepackage[export]{adjustbox}
\newtheorem{theorem}{Theorem}[section]

\newtheorem{remark}[theorem]{Remark}

\def\letters{a,b,c,d,e,f,g,h,i,j,k,l,m,n,o,p,q,r,s,t,u,v,w,x,y,z}
\def\Letters{A,B,C,D,E,F,G,H,I,J,K,L,M,N,O,P,Q,R,S,T,U,V,W,X,Y,Z}
\makeatletter
\@for \@l:=\Letters \do{%
  \expandafter\edef\csname\@l bb\endcsname{\noexpand\ensuremath{\noexpand\mathbb{\@l}}}%
  \expandafter\edef\csname\@l bf\endcsname{{\noexpand\bf \@l}}%
  \expandafter\edef\csname\@l cal\endcsname{\noexpand\ensuremath{\noexpand\mathcal{\@l}}}%
  \expandafter\edef\csname\@l eu\endcsname{\noexpand\ensuremath{\noexpand\EuScript{\@l}}}%
  \expandafter\edef\csname\@l frak\endcsname{\noexpand\ensuremath{\noexpand\mathfrak{\@l}}}%
  \expandafter\edef\csname\@l rm\endcsname{{\noexpand\rm \@l}}%
  \expandafter\edef\csname\@l scr\endcsname{\noexpand\ensuremath{\noexpand\mathscr{\@l}}}%
}
\@for \@l:=\letters \do{%
  \expandafter\edef\csname\@l bf\endcsname{{\noexpand\bf \@l}}%
  \expandafter\edef\csname\@l frak\endcsname{\noexpand\ensuremath{\noexpand\mathfrak{\@l}}}%
  \expandafter\edef\csname\@l scr\endcsname{\noexpand\ensuremath{\noexpand\mathscr{\@l}}}%
}
\makeatother
 \definecolor{shadecolor}{rgb}{0.6, 0.6, 0.6} 
  \definecolor{red}{rgb}{1,0,0} 
  \definecolor{darkgreen}{rgb}{0, 0.6, 0}
\newcommand{\isdef}{\mathrel{\mathrel{\mathop:}=}}

\begin{document}
\title[Samplet basis pursuit]{Samplet basis pursuit: Multiresolution scattered data
approximation with sparsity constraints}
\author{Davide Baroli}
\address{
Davide Baroli,
Istituto Eulero,
Universit\`a della Svizzera italiana, Lugano,
Via la Santa 1, 6962 Lugano, Svizzera.}
\email{michael.multerer@usi.ch}
\author{Helmut Harbrecht}
\address{Helmut Harbrecht,
Departement f\"ur Mathematik und Informatik, 
Universit\"at Basel, 
Spiegelgasse 1, 4051 Basel, Switzerland.}
\email{helmut.harbrecht@unibas.ch}
\author{Michael Multerer}
\address{
Michael Multerer,
Istituto Eulero,
Universit\`a della Svizzera italiana,
Via la Santa 1, 6962 Lugano, Svizzera.}
\email{michael.multerer@usi.ch}
\begin{abstract}
We consider scattered data approximation in samplet 
coordinates with $\ell_1$-re\-gu\-la\-ri\-za\-tion. The application of an 
$\ell_1$-regularization term enforces sparsity of the coefficients 
with respect to the samplet basis.
Samplets are wavelet-type signed measures, 
which are tailored to scattered data. 
Therefore, samplets enable the use of well-established multiresolution 
techniques on general scattered data sets.
They provide similar properties 
as wavelets in terms of localization, multiresolution analysis, and data 
compression. By using the Riesz isometry, we embed samplets 
into reproducing kernel Hilbert spaces and discuss the properties of the 
resulting functions. We argue that the class of signals that are sparse 
with respect to the embedded samplet basis is considerably 
larger than the class of signals that are sparse with respect to the
basis of kernel translates. Vice versa, every signal that is a linear 
combination of only a few kernel translates is sparse in samplet coordinates. 

We propose 
the rapid solution of the problem under consideration by combining 
soft-shrinkage with the semi-smooth Newton method. Leveraging on the sparse 
representation of kernel matrices in samplet coordinates, this approach 
converges faster than the fast iterative shrinkage thresholding algorithm 
and is feasible for large-scale data. Numerical benchmarks are presented 
and demonstrate the superiority of the multiresolution approach over the 
single-scale approach. As large-scale applications, the surface reconstruction 
from scattered data and the reconstruction of scattered temperature data using a dictionary 
of multiple kernels are considered.
\end{abstract}

\maketitle

\section{Introduction}\label{sec:intro}
Sparsity constraints have a wide applicability in machine learning, 
statistics, as well as in signal processing. Examples for the latter
are deblurring, feature selection and 
compressive sensing, see \cite{Candes,BasisPursuit,Donoho,learning,Tao}.
In practice, sparsity constraints are imposed by adding an 
$\ell^1$-regularization term to the objective function. However, sparsity 
constraints only make sense if a basis is used for the representation 
where the data actually becomes sparse. In the past, mostly wavelet bases, 
Fourier bases, or frames like curvelets, contourlets, and shearlets have 
been used as they are known to give rise to sparse representations, 
see \cite{ISTA,curvelets,contourlets,rauhut,shearlets} for instance. 
However, such discretization concepts are based on regular grids 
and it is not obvious how to extend them to scattered data as 
they typically appear in applications.

In the present article, we employ \emph{samplets} 
for imposing sparsity. Samplets have been invented in 
\cite{HM22} and extend the concept of multiresolution 
analyses and respective wavelet bases to scattered
data. The construction relies on the idea of Tausch and 
White \cite{TW03}, who defined orthonormal multiresolution
bases with vanishing moments bottom-up by employing a
hierarchical cluster tree. Indeed, as shown in \cite{HM22},
samplet expansions of smooth functions are sparse, 
allowing for data compression and singularity extraction 
\cite{MH}. As a consequence, kernel matrices issuing
from asymptotically smooth kernels can be compressed 
in samplet representation to very sparse matrices. According 
to \cite{HMSS22}, such samplet compressed kernel 
matrices form a sparse matrix algebra, enabling 
efficient scattered data analysis.

In this article, we consider scattered data approximation, 
which is known to be an ill-posed but uniquely solvable problem, see 
\cite{Fasshauer2007,Wendland2004} for instance. 
By means of the Riesz isometry, we 
embed samplets into reproducing kernel Hilbert 
spaces and examine the properties of the resulting Riesz 
representers. Then, every signal that is a linear combination of 
only a few kernel translates is sparse in samplet coordinates.
However, the class of signals that are sparse with respect 
to the samplets' Riesz representers is much larger than the 
class of signals that are sparse with respect to the basis of 
kernel translates. Thus, in view of the hierachical structure
of the samplet basis, well-established multiresolution techniques 
become applicable on general scattered data sets.

To enforce
sparsity, we apply regularization with respect to the 
$\ell^1$-norm, which improves the feature selection in comparison 
with the standard Tikhonov regularization.
Dealing with such \emph{sparsity constraints} in an efficient 
way is also mandatory for \emph{basis pursuit}, that is, for decomposing 
given data into an optimal superposition of dictionary elements, 
where optimal means having the smallest $\ell^1$-norm of 
coefficients among all such decompositions, see \cite{CDS98,
MZ,tropp} for instance. We demonstrate that the basis pursuit 
problem can efficiently be solved by using a samplet basis. 
In particular, the samplet representation we obtain 
is extremely sparse.

In literature, the $\ell^1$-regularization term is treated 
numerically by using the iterative shrinkage-thresholding 
algorithm (ISTA), see \cite{ISTA}, and its accelerated counterpart 
which is the fast iterative-shrinkage thresholding algorithm (FISTA), 
see \cite{FISTA}. Another approach to solve the underlying 
optimization problem is the alternating direction method 
of multipliers (ADMM), see \cite{ADMM}, which we however do 
not consider here. Instead, we follow the idea of \cite{LorenzGriesse} 
and apply the semi-smooth Newton method, which corresponds
to an active set method, cp.~\cite{CZQ,HIK}. As we will see, this 
leads to an efficient solver for the problem under consideration.
Extensive numerical tests are presented 
and validate the feasibility and power of the proposed
approach. Our benchmarks demonstrate the superiority 
of the multiresolution approach over the single-scale
approach. As large-scale applications, we consider the 
surface reconstruction from measurements of the signed distance function
and the reconstruction of temperature data using 
a dictionary of multiple kernels.

The rest of this article is structured as follows. In 
Section~\ref{section:Samplets}, we introduce
reproducing kernel Hilbert spaces and the scattered 
data approximation problem. Then, in Section~\ref{sec:samplets},
we recapitulate the concept of samplets and introduce 
their embedding into reproducing kernel Hilbert spaces by means of 
the Riesz isometry. In
particular, we discuss the properties of the dual basis and its
role in the sparse representation of scattered data.
Section~\ref{sec:learning} starts from traditional 
kernel ridge regression and then presents scattered data approximation with 
sparsity constraints in samplet coordinates. 
We call the resulting approach samplet basis 
pursuit which especially allows for the usage of a dictionary of 
several kernels. Solution algorithms to compute
sparse samplet expansions, particularly suited for
large-scale data, are proposed in Section~\ref{sec:algos}. 
Extensive numerical tests are presented in Section~\ref{sec:results}. 
Finally, in Section \ref{sct:conclusion}, we draw
the conclusion of the article.

\section{Problem formulation}\label{section:Samplets}
Let \((\Hcal,\langle\cdot,\cdot\rangle_\Hcal)\) be a 
Hilbert space of functions \(h\colon\Omega\to\Rbb\)
for some set \(\Omega\subset\Rbb^d\). Furthermore, 
let \(\kernel\) be a continuous, symmetric, and positive 
definite kernel, i.e., for any set \(X\isdef\{{\bm x}_1,
\ldots,{\bm x}_N\}\subset\Omega\) of mutually distinct points, 
the \emph{kernel matrix} 
\begin{equation}\label{eq:KernelMatrix}
{\bm K}\isdef [\kernel({\bm x}_i,{\bm x}_j)]_{i,j=1}^N\in\Rbb^{N\times N}
\end{equation}
is symmetric and positive definite. The kernel \(\kernel\) is the 
reproducing kernel of \(\Hcal\), iff \(\kernel({\bm x},\cdot)\in\Hcal\) 
for every \({\bm x}\in\Omega\) and \(h({\bm x})=\langle\kernel
({\bm x},\cdot),h\rangle_\Hcal\) for every \(h\in\Hcal\). In this 
case, we call \((\Hcal,\langle\cdot,\cdot\rangle_\Hcal)\) a 
\emph{reproducing kernel Hilbert space}.

For a continuous function $f\in C(\Omega)$, 
we shall use the notation
\[
(f,\delta_{{\bm x}})_\Omega\isdef 
\delta_{{\bm x}}(f)=f({\bm x}),\quad{\bm x}\in\Omega,
\]
for the point evaluation, i.e.,
$\delta_{\bm x}\in[C(\Omega)]'$ is the point evaluation functional
at ${\bm x}\in\Omega$. Since the kernel 
\(\kernel({\bm x},\cdot)\) is the Riesz representer of
the point evaluation \((\cdot,\delta_{\bm x})_\Omega\), 
we particularly have 
\begin{equation}\label{eq:Repprop}
(h,\delta_{\bm x})_\Omega
=\langle\kernel({\bm x},\cdot),h\rangle_\Hcal\quad\text{for every }
h\in\Hcal.
\end{equation}
Given the set 
\(X=\{{\bm x}_1,\ldots,{\bm x}_N\}\subset\Omega\),
we introduce the subspace
\begin{equation}\label{eq:HX}
\begin{aligned}
\Hcal_X\isdef
&\spn\{\phi_1,\ldots,\phi_N\}
\subset\Hcal,\\
&\qquad\quad\phi_i\isdef\kernel({\bm x}_i,\cdot)\text{ for }
i=1,\ldots,N.
\end{aligned}
\end{equation}
We call the basis \(\phi_1,\ldots,\phi_N\) of \(\Hcal_X\)
the 
\emph{basis of kernel translates}. 

The subspace \(\Hcal_X\) is isometrically
isomorphic to the subspace \(\Xcal\isdef\spn\{\delta_{{\bm x}_1},\ldots,
\delta_{{\bm x}_N}\}\subset\Hcal'\) 
by means of the Riesz isometry. 
We may thus identify
\[
u'=\sum_{i=1}^Nu_i\delta_{{\bm x}_i}\in\Xcal\quad\text{with}\quad
u=\sum_{i=1}^Nu_i\kernel({\bm x}_i,\cdot)\in\Hcal_X.
\]

To provide a notion of orthogonality in \(\Xcal\), we introduce
the inner product 
\begin{equation*}
\langle u',v'\rangle_{\Xcal}\isdef\sum_{i=1}^N u_iv_i,\quad
u'=\sum_{i=1}^Nu_i\delta_{{\bm x}_i},\ v'=\sum_{i=1}^Nv_i\delta_{{\bm x}_i}.
\end{equation*}
We remark that this inner product differs from the
restriction of the canonical one in \(\Hcal\) to \(\Hcal_X\).
The latter is given by
\[
\langle u,v\rangle_\Hcal={\bm u}^\intercal{\bm K}{\bm v}
\]
with the kernel matrix \({\bm K}\) from \eqref{eq:KernelMatrix}, and
the coefficient vectors
\({\bm u}\isdef[u_i]_{i=1}^N\), \({\bm v}\isdef[v_i]_{i=1}^N\).

A direct consequence of the duality between \(\Hcal_X\)
and \(\Xcal\) is that
the \(\Hcal\)-orthogonal projection
of a function \(h\in\Hcal\) onto \(\Hcal_X\) is given by the 
interpolant
\begin{equation*}
s_h\isdef\sum_{i=1}^N\alpha_i\kernel({\bm x}_i,\cdot),
\end{equation*}
which satisfies 
\begin{equation}\label{eq:interpolation}
s_h({\bm x}_i) = h({\bm x}_i)\quad\text{for all }{\bm x}_i\in X.
\end{equation}
In other words, the interpolation problem \eqref{eq:interpolation}
corresponds to a 
Galerkin formulation for the \(\Hcal\)-orthogonal projection 
onto \(\Hcal_X\). There holds \eqref{eq:interpolation}, iff
\begin{equation}\label{eq:Galerkin}
\langle s_h,v\rangle_\Hcal
=\langle h,v\rangle_\Hcal\quad\text{for all }v\in\Hcal_X.
\end{equation}

Choosing the basis of kernel translates \(\phi_i=\kernel({\bs x}_i,\cdot)\)
as ansatz- and test functions in \eqref{eq:Galerkin},
the expansion coefficients 
\[{\bm\alpha}\isdef[\alpha_i]_{i=1}^N
\]
can be retrieved by solving the linear system
\begin{equation}\label{eq:LSE}
{\bm K}{\bm\alpha}={\bm h},\quad{\bm h}\isdef[h({\bm x}_i)]_{i=1}^N.
\end{equation}
Depending on the choice of the kernel function, 
the linear system \eqref{eq:LSE}
is typically ill-conditioned and a suitable regularization
is required to obtain a solution. Within this article, our
focus is on \(\ell^1\)-regularization with respect to an
appropriate multiresolution analysis. The particular
multiresolution basis is introduced in the next section.

\section{Samplets}\label{sec:samplets}
\subsection{Samplet bases}
We equip the space \(\Xcal\subset\Hcal'\) with a multiresolution analysis
\[
\Scal_j=\spn\{\sigma_{j,k}\}_k,\quad j=0,\ldots J,
\] 
whose basis elements are called \emph{samplets},
such that
\[
\Xcal=\Xcal_J=\bigoplus_{j=0}^J\Scal_j\quad\text{and}\quad
\Scal_{\ell}\perp\Scal_{j}\text{ for }\ell\neq j.
\] 
Each samplet has a representation with respect
to the point evaluation functionals in \(\Xcal\), i.e.,
\begin{equation}\label{eq:sampletRep}
\sigma_{j,k}=\sum_{i=1}^n\omega_{j,k,i}\delta_{{\bm x}_i}.
\end{equation} 
For the expansion coefficients, we introduce the
coefficient vectors
${\bm\omega}_{j,k}=\big[\omega_{j,k,i}\big]_i$.
In view of \eqref{eq:sampletRep},
samplets can be defined on arbitrary scattered data
sets. In particular, the samplet basis can be constructed in linear 
time with respect to the number of points $N$ in a black-box fashion 
and in arbitrary dimension. In this regard, 
they overcome well known limitations of wavelets, which are
usually constructed only in one spatial dimension on equispaced
data sets.
For all the details on the construction
samplets, we refer to \cite{HM22}.

The properties of samplets are summarized by the following theorem, 
which is a compilation of results from \cite{HKS05,TW03,HM22}
for cardinality balanced hierarchical cluster trees for \(X\).
Note that the \emph{support} \(\nu_{j,k}\) of a samplet \(\sigma_{j,k}\) 
is to be understood as the smallest convex set which contains 
all points with non-vanishing samplet coefficients.

\begin{theorem}\label{theo:waveletProperties}
The samplet basis \(\bigcup_{j=0}^J\{\sigma_{j,k}\}_k\)
forms an orthonormal basis in $\Xcal$, satisfying
the following properties:
\begin{enumerate}
\item 
There holds $c^{-1} 2^j\leq\operatorname{dim}\Scal_j\leq c 2^j$
for a constant \(1<c\leq 2\).
\item 
The samplets have vanishing moments of order $q+1$, 
i.e., 
\((p,\sigma_{j,k})_\Omega = 0\) for all \(p\in\Pcal_q(\Omega)\),
 where \(\Pcal_q(\Omega)\) is the space of polynomials up 
 to degree \(q\).
\item 
The coefficient vector 
${\bm\omega}_{j,k}=\big[\omega_{j,k,i}\big]_i$ 
of the samplet $\sigma_{j,k}$ satisfies 
\(\|{\bm\omega}_{j,k}\|_{1}\leq c 2^{(J-j)/2}\) for a
constant \(0<c<2\).
\item Given $f\in C^{q+1}(\nu_{j,k})$, there holds 
\begin{equation*}
\begin{aligned}
&|(f,\sigma_{j,k})_\Omega|\\
&\qquad\le \bigg(\frac{d}{2}\bigg)^{q+1}
  	\frac{(\operatorname{diam}\nu_{j,k})^{q+1}}{(q+1)!}
	\|f\|_{C^{q+1}(\nu_{j,k})}\|{\bm\omega}_{j,k}\|_{1}.
	\end{aligned}
\end{equation*}
\end{enumerate}
\end{theorem}
Property 4) of Theorem~\ref{theo:waveletProperties} 
is crucial when it comes to feature detection. Regions
where the data is smooth will lead to negligible expansion
coefficients, while there is no decay with respect to the
scale in regions where the data is non-smooth.

The transformation of a functional \(u'\in\Xcal\) into its
samplet representation can be performed with linear cost using
the fast samplet transform, compare \cite{HM22}.
We denote this basis transform in the following by
\[
u'=\sum_{i=1}^N u_i\delta_{{\bs x}_i}=
\sum_{i=1}^N [{\bs T}{\bs u}]_i\sigma_i,
\]
with a suitable linear ordering \(i=i(j,k)\) of the samplets,
for example a breadth-first-search-like ordering. 
Furthermore \({\bs T}\in\Rbb^{N\times N}\) 
denotes the orthogonal samplet transformation matrix, i.e.,
\(
{\bs T}{\bs T}^\intercal={\bs T}^\intercal{\bs T}={\bs I}.
\)

\subsection{Samplets in reproducing kernel Hilbert spaces}
The samplet basis gives rise to a multiresolution 
basis in \(\Hcal_X\) by considering the embedding
\begin{equation}\label{eq:Wavelet-Kern}
\sigma_{j,k}=\sum_{i=1}^N\omega_{j,k,i}\delta_{{\bs x}_i}\mapsto
\psi_{j,k}=\sum_{i=1}^N\omega_{j,k,i}\kernel({\bs x}_i,\cdot)
\end{equation}
by means of the Riesz isometry, compare
\eqref{eq:Repprop}.
Especially, for \(j>0\) there holds
\(
\langle\psi_{j,k},h\rangle_\Hcal=0
\)
for any \(h\in\Hcal\) which satisfies
\(h|_{\nu_{j,k}}\in\Pcal_q(\nu_{j,k})\), which
means that the function \(\psi_{j,k}\) has vanishing moments
in \(\Hcal\). 
Defining 
\[\Wcal_j\isdef\operatorname{span}\{\psi_{j,k}\}_k,
\]
we obtain the primal multiresolution analysis
\[
\Hcal_X=\bigoplus_{j=0}^J\Wcal_j.
\]
Using the functions \(\psi_{j,k}\) as ansatz-
and test functions in the Galerkin formulation \eqref{eq:Galerkin}
yields the linear system
\begin{equation}\label{eq:Tlinsys}
{\bs K}^\Sigma{\bs\beta}\isdef{\bs T}{\bs K}{\bs T}^\intercal\bs\beta
={\bs T}{\bs h},
\end{equation}
i.e.,
\[
[\langle\psi_{j,k},\psi_{\ell,m}\rangle_\Hcal]_{j,k,\ell,m}=
{\bs T}{\bs K}{\bs T}^\intercal\]
and
\[
[\langle\psi_{j,k},h\rangle_\Hcal]_{j,k}={\bs T}{\bs h}.
\]
The solution of the linear system \eqref{eq:Tlinsys} 
is equivalent to the one of \eqref{eq:LSE} by the transform
\begin{equation}\label{eq:coeffTrafo}
{\bs\beta}={\bs T}{\bs\alpha}={\bs T}{\bs K}^{-1}{\bs h}.
\end{equation}

Note that a compressed version \({\bs K}_\varepsilon^\Sigma\) 
of the transformed kernel matrix \({\bs K}^\Sigma\) can
efficiently be computed by combining the compression 
with a hierarchical matrix approach. Indeed, 
if the set $X$ is quasi-uniform, i.e., separation radius and
fill-distance are of comparable size, then the matrix
\({\bs K}_\varepsilon^\Sigma\) which is compressed with a 
relative compression error smaller than \(\varepsilon\) 
in the Frobenius norm has only \(\Ocal(N\log N)\) nonzero 
matrix entries and the cost for its computation is
\(\Ocal(N\log N)\).

\subsection{Characterization of the dual basis}\label{ss:biosamplets}
Letting 
\[
\widetilde{\bs\omega}_{j,k}\isdef{\bs K}^{-1}{\bs\omega}_{j,k},\quad
\widetilde{\sigma}_{j,k}\isdef\sum_{i=1}^N\widetilde{\omega}_{j,k,i}\delta_{{\bs x}_i},
\]
the corresponding \emph{biorthogonal (samplet) basis}, also called 
\emph{dual basis}, is given by
\[
\widetilde{\psi}_{j,k}=\sum_{i=1}^N\widetilde{\omega}_{j,k,i}\kernel({\bs x}_i,\cdot).
\]
The dual basis satisfies indeed
\[
\begin{aligned}
\big\langle{\psi}_{j,k},\widetilde{\psi}_{\ell,m}\big\rangle_\Hcal
&={\bs\omega}_{j,k}^\intercal{\bs K}\widetilde{\bs\omega}_{\ell,m}
={\bs\omega}_{j,k}^\intercal{\bs K}{\bs K}^{-1}{\bs\omega}_{\ell,m}\\
&={\bs\omega}_{j,k}^\intercal{\bs\omega}_{\ell,m}=\delta_{j,\ell}\delta_{k,m}
\end{aligned}
\]
as required by the definition. Therefore, 
defining
\[
\widetilde{\Wcal}_j\isdef\operatorname{span}\{\widetilde{\psi}_{j,k}\}_k
\]
yields the dual multiresolution analysis
\[
\Hcal_X=\bigoplus_{j=0}^J\widetilde{\Wcal}_j,\quad
{\Wcal}_j\perp\widetilde{\Wcal}_\ell\quad\text{for }j\neq\ell.
\]

\begin{remark}
Since \(\kernel\) is particularly a Mercer kernel, see \cite{Koenig}
for instance, it exhibits an \(L^2(\Omega\times\Omega)\)-convergent 
spectral representation
\[
\kernel({\bs x},{\bs y})=\sum_{i}\lambda_i k_i({\bs x})k_i({\bs y})
\]
with \((k_i,k_j)_{L^2(\Omega)}=\delta_{i,j}\) as well as 
\(\langle\hat{k}_i,\hat{k}_j\rangle_{\Hcal}=\delta_{i,j}\), 
where we set \(\hat{k}_i\isdef\sqrt{\lambda_i}k_i\). In the 
spectral basis, the primal and dual basis functions 
can be written as
\begin{align*}
\psi_{j,k}&=\sum_{i}(\hat{k}_i,\sigma_{j,k})_\Omega\hat{k}_i\in\Hcal_X,\\
\widetilde{\psi}_{j,k}&=\sum_{i}(\hat{k}_i,\widetilde{\sigma}_{j,k})_\Omega\hat{k}_i\in\Hcal_X
\end{align*}
for all $j\leq J$.
\end{remark}

The dual basis gives rise to the representations
\[
h=\sum_{i=1}^N\big\langle\widetilde{\psi}_{i},h\big\rangle_\Hcal\psi_i
=\sum_{i=1}^N\big\langle{\psi}_{i},h\big\rangle_\Hcal\widetilde{\psi}_i
\quad\text{for all }h\in\Hcal_X.
\]
Especially, employing the dual basis, the solution to
\eqref{eq:Galerkin} can directly be represented according to
\begin{equation}\label{eq:waveletInterpolant}
s_h=\sum_{i=1}^N\beta_i\psi_i=\sum_{i=1}^N
\big\langle\widetilde{\psi}_i,h\big\rangle_\Hcal\psi_i.
\end{equation}

\subsection{Sparse kernel interpolants}
Straightforward calculation yields the identities
\[
\big\langle{\psi}_{j,k},h\big\rangle_\Hcal
={\bs\omega}_{j,k}^\intercal{\bs h}\]
and
\[
\big\langle\widetilde{\psi}_{j,k},h\big\rangle_\Hcal
={\bs\omega}_{j,k}^\intercal{\bs K}^{-1}{\bs h}
\]
for any $h\in\Hcal$.
Consequently, there holds
\[
\big\langle\widetilde{\psi}_{j,k},h\big\rangle_\Hcal=0
\]
for any \(h\in\Hcal_X\) which satisfies
\(h=\sum_{i=1}^N\alpha_i k({\bs x}_i,\cdot)\) and \(\alpha_i=p({\bs x}_i)\), where
\[p|_{\nu_{j,k}}\in\Pcal_q(\nu_{j,k}).\] The latter means
that the coefficient vector looks like a polynomial
evaluated at the data sites \({\bs x}_i\in\nu_{j,k}\). 
This implies that piecewise smooth coefficient vectors 
amount to sparse representations with respect to the basis 
\(\psi_1,\ldots,\psi_N\). In addition, in view of \eqref{eq:coeffTrafo}, 
sparse coefficient vectors with respect to the basis \(\phi_1,\ldots,\phi_N\) 
give rise to sparse coefficient vectors in the basis \(\psi_1,\ldots,\psi_N\) 
due to the locality of the samplet supports. Therefore, we conclude that
sparsity in the samplet basis is the more general concept 
compared to the single-scale basis.

Dependent on the kernel \(\kernel\) under consideration, or the 
specific application, the positive definiteness assumption of the 
kernel matrix $\bs K$ can numerically be violated. In such cases,
a suitable regularization is required. Two common approaches, tailored
to the samplet framework, are presented in the following section.

\section{Scattered data approximation}\label{sec:learning}
\subsection{Kernel ridge regression}
We first comment on the traditional kernel ridge regression.
Here, one considers the 
squared loss function with ridge regularization in \(\Hcal\) which 
yields the optimization problem
\[
\min_{s_h\in\Hcal_X}\frac 1 2\sum_{i=1}^N\big|s_h({\bs x}_i)-h({\bs x}_i)\big|^2
+\frac{\lambda}{2}\|s_h\|_\Hcal^2.
\]
This problem can be reformulated in matrix-vector form as
\[
\min_{{\bs\alpha}\in\Rbb^N}\frac 1 2\|{\bs h}-{\bs K}{\bs\alpha}\|_2^2
+\frac{\lambda}{2}{\bs\alpha}^\intercal{\bs K}{\bs\alpha}.
\]
Straightforward calculation yields the first order optimality condition
\begin{equation}\label{eq:KRR}
({\bs K}+\lambda{\bs I}){\bs\alpha}={\bs h}\quad\text{or}\quad
({\bs K}^\Sigma+\lambda{\bs I}){\bs\beta}={\bs T}{\bs h},
\end{equation}
due to the unitary invariance of the euclidean norm, compare
\eqref{eq:LSE} and \eqref{eq:Tlinsys}. The 
first expression computes the coefficients $\bs\alpha\in\mathbb{R}^N$
of the kernel interpolant with respect to the single-scale basis 
while the second expression leads to the respective representation 
in samplet coordinates, i.e.,
\[
  s_h({\bs x}) = \sum_{i=1}^N\alpha_i\kernel({\bm x}_i,{\bs x})=
  \sum_{i=1}^N\alpha_i\phi_i({\bm x})
  	= \sum_{i=1}^N\beta_i\psi_i({\bs x}),
\]
compare \eqref{eq:Wavelet-Kern}. In particular, we observe that the
regularization term is invariant under the samplet transform.

The smallest eigenvalue and hence the condition number of the systems
in \eqref{eq:KRR} is steered by the regularization parameter \(\lambda\).
The matrix \({\bs K}+\lambda{\bs I}\) is strictly positive definite 
for \(\lambda>0\) independent of the size $N$ and, therefore, 
the conjugate gradient method can be used to solve the system 
of linear equations. The required matrix-vector multiplications 
can efficiently be realized in samplet coordinates. Depending 
on the condition number of the regularized system, the
convergence might deteriorate and preconditioning becomes an issue.
Diagonal scaling can be used to mitigate the ill-conditioning and seems
to work in our numerical tests. However, we stress that the current 
samplet construction does not provide sufficient norm equivalences 
to render diagonal scaling a provable preconditioner. Alternatively, 
a (partial) sparse inverse is efficiently computable, cf.\ 
\cite{HM22,HMSS22}, and can serve as an algebraic 
preconditioner.
\subsection{Samplet basis pursuit}
In this subsection, we 
discuss \(\ell^1\)-regularization which is known to 
promote sparsity, compare \cite{Candes,Donoho,ISTA}. With regard 
to the basis of kernel translates, we consider the functional
\begin{equation}\label{eq:SSLASSO}
\min_{{\bs\alpha}\in\Rbb^N}\frac 1 2\|{\bs h}-{\bs K}{\bs\alpha}\|_2^2
+\sum_{i=1}^N w_i|\alpha_i|
\end{equation}
for a given weight vector \({\bs w}\in\Rbb^N\), see \cite{LorenzGriesse}.

As has been argued in Subsection~\ref{ss:biosamplets}, a wider class of
sparse solutions can be retrieved by using a samplet basis. This means
that we consider
\begin{equation}\label{eq:MSLASSO}
\min_{{\bs\alpha}\in\Rbb^N}\frac 1 2\|{\bs h}-{\bs K}{\bs\alpha}\|_2^2
+\sum_{i=1}^N w_i|\beta_i|,\quad\text{where }{\bs\beta}={\bs T}{\bs\alpha}.
\end{equation}
In what follows, we shall
refer to \eqref{eq:MSLASSO} as \emph{samplet basis pursuit}.

We remark that, in the spirit of \cite{GCG05,CDS98}, also a dictionary
of multiple kernels can be employed in \eqref{eq:SSLASSO} and 
\eqref{eq:MSLASSO}. Given kernels $\kernel_1,
\ldots,\kernel_L$, we are then looking for a sparse representation 
of the form
\[
s_h=\sum_{j=1}^L\sum_{i=1}^N\alpha_{i}^{({j})}\kernel_j({\bs x}_i,\cdot)
=\sum_{j=1}^L\sum_{i=1}^N\beta_{i}^{({j})}\psi_{i}^{(j)}({\bs x}),
\]
compare \eqref{eq:waveletInterpolant}. Setting 
\[
\bs K\isdef[{\bs K}_1,\ldots,{\bs K}_L],\ {\bs K}_j\isdef[\kernel_j({\bs x}_i,{\bs x}_j)]_{i,j=1}^N,
\] and 
\[
\bs\alpha^\intercal
\isdef[{\bs\alpha}_{1}^\intercal,\ldots,{\bs\alpha}_{L}^\intercal],\ {\bs\alpha}_j\in\Rbb^N,
\]
this approach also amounts to \eqref{eq:SSLASSO} and
\eqref{eq:MSLASSO} with the obvious modifications.
The most important difference
to the original problem is, of course, that the matrix $\bs K$ is 
not quadratic any more which means that the underlying linear system 
of equations $\bs K\bs\alpha = \bs h$ is underdetermined.

Note that the weight vector 
${\bs w}\isdef[w_i]_i$ 
in \eqref{eq:SSLASSO} and
\eqref{eq:MSLASSO}, respectively, plays the role of the 
regularization parameter, where each coefficient is regularized 
individually. We refer to, e.g., \cite{ISTA,Lorenz,RT} for the 
analysis of the regularizing properties and for appropriate 
parameter choice rules. Numerical algorithms to solve 
the optimization problems \eqref{eq:SSLASSO} and
\eqref{eq:MSLASSO} are based on soft-thresholding. 
We recapitulate two efficient numerical realizations in 
the subsequent Section~\ref{sec:algos}.

\section{Algorithms for samplet basis pursuit}\label{sec:algos}
\subsection{Fast iterative-shrinkage thresholding algorithm}
Leveraging on the fast matrix-vector product in samplet coordinates,
problem \eqref{eq:SSLASSO} can be solved by using the fast iterative 
shrinkage-thresholding algorithm (FISTA) from \cite{FISTA}. Departing
from the original algorithm, two samplet transforms are applied in each 
step to exploit the samplet matrix compression. The resulting algorithm 
can be found in Algorithm~\ref{algo:FISTA}. 
In this algorithm, and in the following ones, we use the soft-shrinkage
operator
\[
\operatorname{SS}_{{\bs w}}({\bs v})
\isdef\operatorname{sign}({\bs v})\max\{{\bs 0},|\bs v|-{\bs w}\},
\]
where all operations have to be understood coordinate-wise. 

Note that we consider 
here the fixed step size version, since the step size \(\delta\), which 
is based on the Lipschitz constant of \({\bs K}^\intercal{\bs K}\), can 
easily be computed by performing a few steps of the power iteration.

\begin{algorithm}
\caption{Fast iterative shrinkage-thresholding algorithm (FISTA)}
\label{algo:FISTA}	
\KwData{Kernel matrix ${\bs K}^\Sigma_\varepsilon$, data ${\bs h}^\Sigma$,
weight ${\bs w}$, initial guess \({\bs\alpha}_0\).}
\KwResult{Coefficient vector ${\bs\alpha}$}	
\Begin{
set ${\bs\gamma}_1\isdef{\bs\alpha}_0$, ${\bs\eta}_1\isdef{\bs T}{\bs\gamma}_1$, $t_1=1$

compute \(\delta\isdef1/\big\|{\bs K}^\Sigma_\varepsilon\big\|_2^2\)

\For{$i=1,2,\ldots$}{
compute \({\bs\alpha}_k\isdef\operatorname{SS}_{\delta{\bs w}}\Big({\bs T}^\intercal\big(
{\bs\eta}_k-\delta({\bs K}^\Sigma_\varepsilon)^\intercal({\bs K}^\Sigma_\varepsilon{\bs\eta}_k
-{\bs h}^\Sigma)\big)\Big)\)

set $\begin{displaystyle}t_{k+1}\isdef\frac{1+\sqrt{1+4t_k^2}}{2}\end{displaystyle}$

set $\begin{displaystyle}{\bs\gamma}_{k+1}\isdef{\bs\alpha}_{k}+\frac{t_k-1}{t_{k+1}}({\bs\alpha}_k
-{\bs\alpha}_{k-1})\end{displaystyle}$

compute \({\bs\eta}_{k+1}\isdef{\bs T}{\bs\gamma}_{k+1}\)
}
}
\end{algorithm}

In case of problem \eqref{eq:MSLASSO}, the FISTA algorithm 
can be employed without any modification by using samplet 
coordinates everywhere. We shall refer to
this variant as MRFISTA. 
%
%
%
%
%

\subsection{Semi-smooth Newton method}
A particularly efficient approach for the solution of \eqref{eq:MSLASSO}
is the semi-smooth Newton method, suggested in \cite{LorenzGriesse},
which is applied to the root finding problem
\[
{\bs 0}=
{\bs\beta}^\star-\operatorname{SS}_{\gamma{\bs w}}\big(
{\bs\beta}^\star+\gamma({\bs K}^{\Sigma})^\intercal({\bs h}^{\Sigma}
-{\bs K}^{\Sigma}{\bs\beta}^\star)\big),\ \gamma>0.
\]
For the reader's convenience, we recall the method in 
Algorithm~\ref{algo:SSN}.
\begin{algorithm}
\caption{Multiresolution semi-smooth Newton method (MRSSN)}
\label{algo:SSN}	
\KwData{Kernel matrix ${\bs K}^\Sigma_\varepsilon$, data ${\bs h}^\Sigma$,
weight ${\bs w}$, initial guess \({\bs\beta}_0\), parameter \(\gamma>0\),
tolerance $\texttt{tol}$.}
\KwResult{Coefficient vector ${\bs\beta}$}	
\Begin{
set ${\bs\beta\isdef{\bs\beta}_0}$ and ${\bs r}\isdef\Fcal({\bs\beta}_0)$

\While{$\|{\bs r}\|>\textnormal{\texttt{tol}}$}{
compute 
\begin{align*}
{\Acal}&\isdef\big\{k\in\{1,\ldots,N\}:\\
&\qquad
\big|[{\bs\beta}+\gamma({\bs K}^\Sigma_\varepsilon)^\intercal({\bs h}^{\Sigma}\!-\!{\bs K}^\Sigma_\varepsilon{\bs\beta})]_k\big|>\gamma w_k \big\}\\
\Ical&\isdef\{1,\ldots,N\}\setminus\Acal
\end{align*}

set \({\bs M}_{\Acal\Acal}\isdef{\bs I}_\Acal({\bs K}^\Sigma_\varepsilon)^\intercal
{\bs K}^\Sigma_\varepsilon{\bs I}_\Acal\) and \({\bs M}_{\Acal\Ical}\isdef
{\bs I}_\Acal({\bs K}^\Sigma_\varepsilon)^\intercal{\bs K}^\Sigma_\varepsilon{\bs I}_\Ical\)

solve \(\gamma{\bs M}_{\Acal\Acal}{\bs\delta} =
\gamma{\bs M}_{\Acal\Ical}{\bs r}-{\bs I}_\Acal{\bs r}\)

set ${\bs\beta}\isdef{\bs\beta}+{\bs\delta}$

set ${\bs r}\isdef\Fcal({\bs\beta})$
}
}
\end{algorithm}

If \(|\Acal|=m\), the matrix \({\bs M}_{\Acal\Acal}\) can be formed with cost
\(\Ocal(m^2N)\). In particular, if the active set is of moderate size,
a direct solver may be employed to solve the corresponding linear system
with cost \(\Ocal(m^3)\).

Due to the local convergence of Newton's method, if the functional under consideration
is numerically not strictly convex, it is favorable to iteratively
decrease the regularization parameter. This procedure has been investigated earlier in
\cite{BNS97} and is summarized in Algorithm~\ref{algo:IRSSN}.

\begin{algorithm}\label{algo:IRMRSSN}
\caption{Iteratively regularized MRSSN}
\label{algo:IRSSN}	
\KwData{Kernel matrix ${\bs K}^\Sigma_\varepsilon$, data ${\bs h}^\Sigma$,
weight ${\bs w}$, initial guess \({\bs\beta}_0\), parameter \(\gamma>0\),
tolerance $\texttt{tol}$.}
\KwResult{Coefficient vector ${\bs\beta}$}	
\Begin{
set ${\bs\beta\isdef{\bs\beta}_0}$ and choose $\mu,\mu_0>1$

\While{$\mu>1$}{
set ${\bs\beta}\isdef\texttt{MRSSN}\big({\bs K}^\Sigma_\varepsilon,{\bs h}^\Sigma,
\eta{\bs w},
{\bs\beta},\gamma,\texttt{tol}\big)$

set $\mu\isdef\min\{1,\mu/\mu_0\}$
}
}
\end{algorithm}

\section{Numerical results}\label{sec:results}
For the numerical experiments in
Subsections~\ref{subs:bench} and \ref{ss:Surf}, we consider a polynomial degree
\(p=6\) for the multipole expansion and samplets with \(q+1=4\) vanishing moments. 
For the experiment in Subsection~\ref{ss:ERA}, we set \(p=4\) and \(q+1=3\).
For the admissibility condition, we choose in all cases \(\eta=d\), where \(d\) 
is the spatial dimension, see \cite{HM22} for a definition of these
parameters.
Moreover, an a-posteriori compression is applied to 
\({\bs K}^\Sigma_\varepsilon\) which means that matrix entries smaller than 
\(10^{-4}\) in modulus are dropped. 
For MRSSN, we consider throughout the numerical experiments the iteratively
regularized version from Algorithm~\ref{algo:IRMRSSN} and set \(\mu_0=1.05\).
The initial parameter is set to \(\mu=\mu_0^{250}\), i.e., we perform 250 iterative 
regularization steps. Moreover, the parameter \(\gamma\) is estimated in a lazy fashion as the smallest
eigenvalue of the matrix \({\bs M}_{\Acal\Acal}\), whenever the cardinality of the
current active set changes by at least 2. The entailed cost is \(\Ocal(m^3)\), where \(|\Acal|=m\),
and is thus of the same order as the solution of the linear system on the active set 
\(\Acal\).

All computations have been performed at
the Centro Svizzero di Calcolo Scientifico (CSCS) on
a single node of the Alps cluster with two AMD EPYC 7742 @2.25GHz
CPUs and up to 512GB of main memory\footnote{
\texttt{https://www.cscs.ch/computers/alps}}.
For the samplet compression, up to 32 cores have been used.
The samplet implementation is available online as software package
\texttt{FMCA}\footnote{\texttt{https://github.com/muchip/fmca}}
and relies on the
\texttt{Eigen} template
library\footnote{\texttt{https://eigen.tuxfamily.org/}}.

\subsection{Benchmarks}\label{subs:bench}
\begin{figure*}[!t]
\begin{center}
\begin{tikzpicture}
\draw(3.7,8)node{\includegraphics[scale=0.06,clip,trim=230 30 220 100]{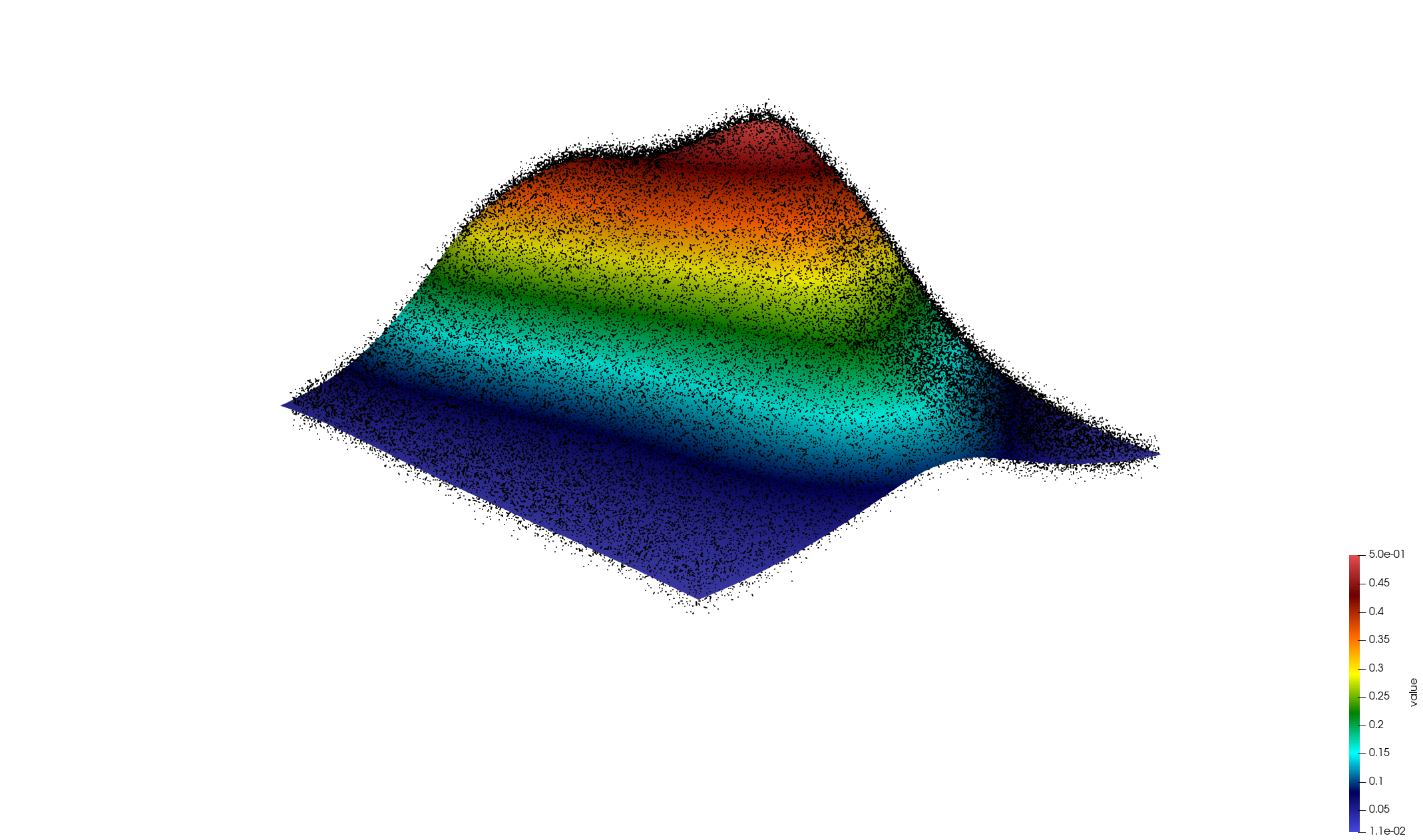}};
\draw(8,8)node{\includegraphics[scale=0.06,clip,trim=230 30 220 100]{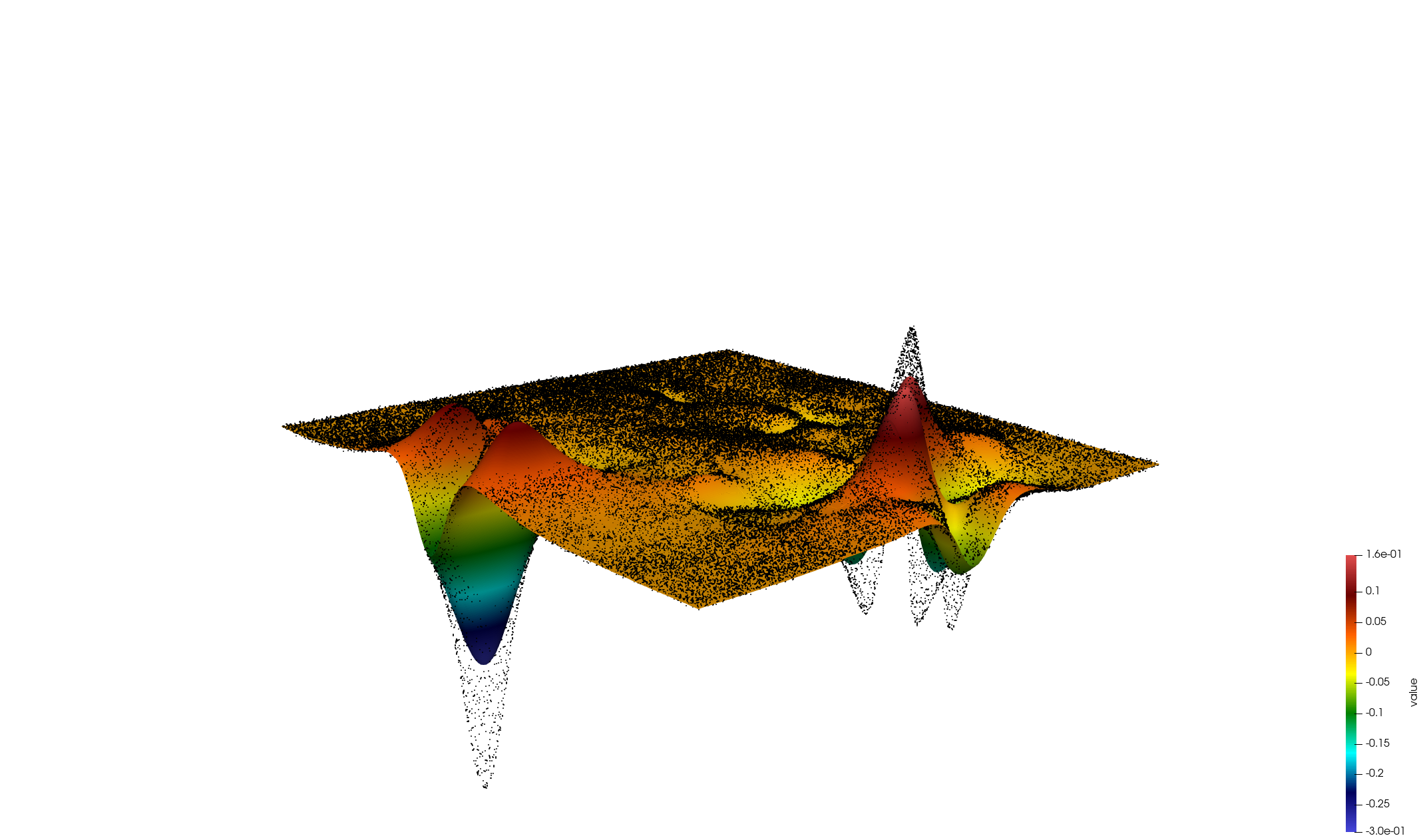}};
\draw(12.3,8)node{\includegraphics[scale=0.06,clip,trim=230 30 220 100]{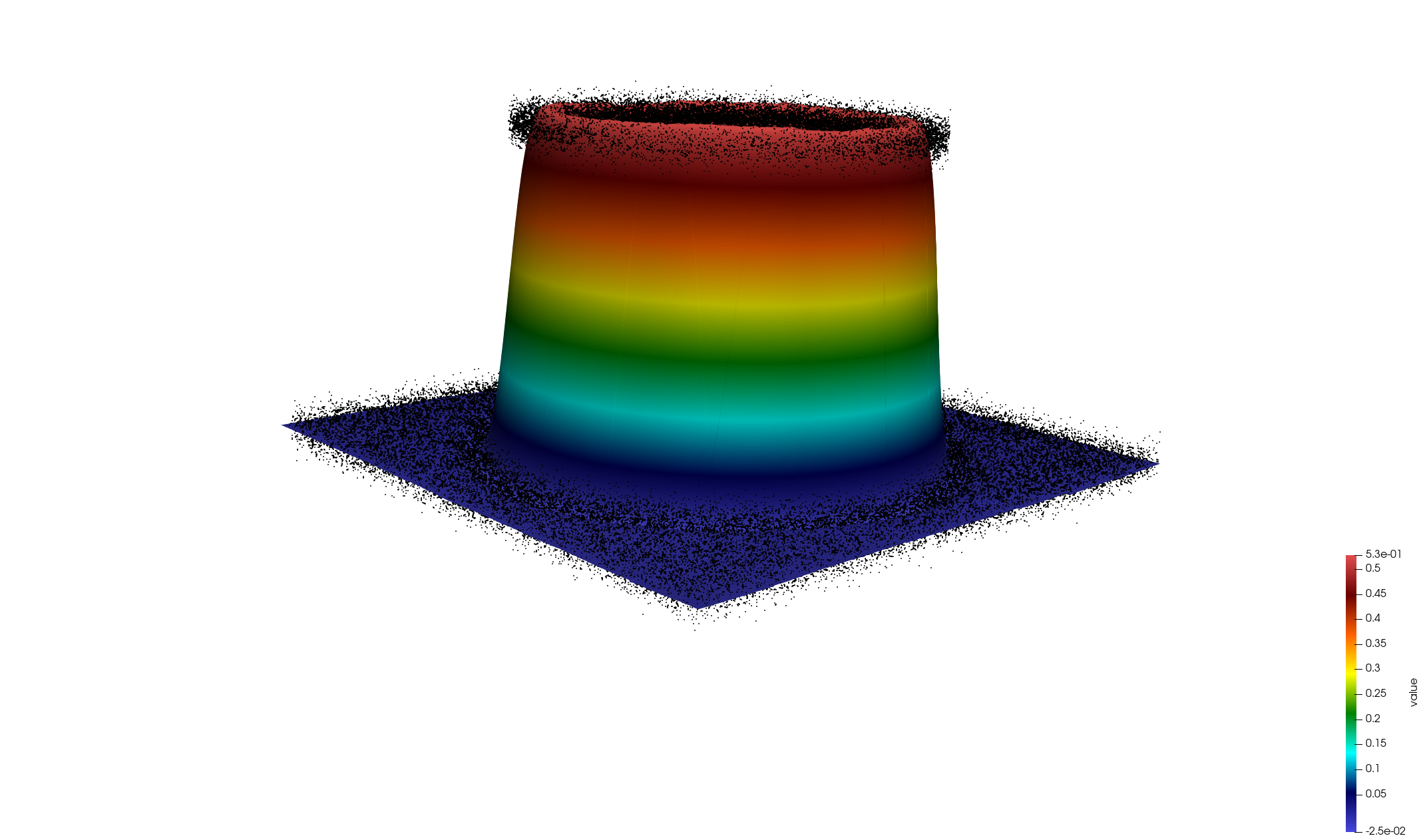}};
\draw(3.7,6)node{\includegraphics[scale=0.06,clip,trim=230 30 220 100]{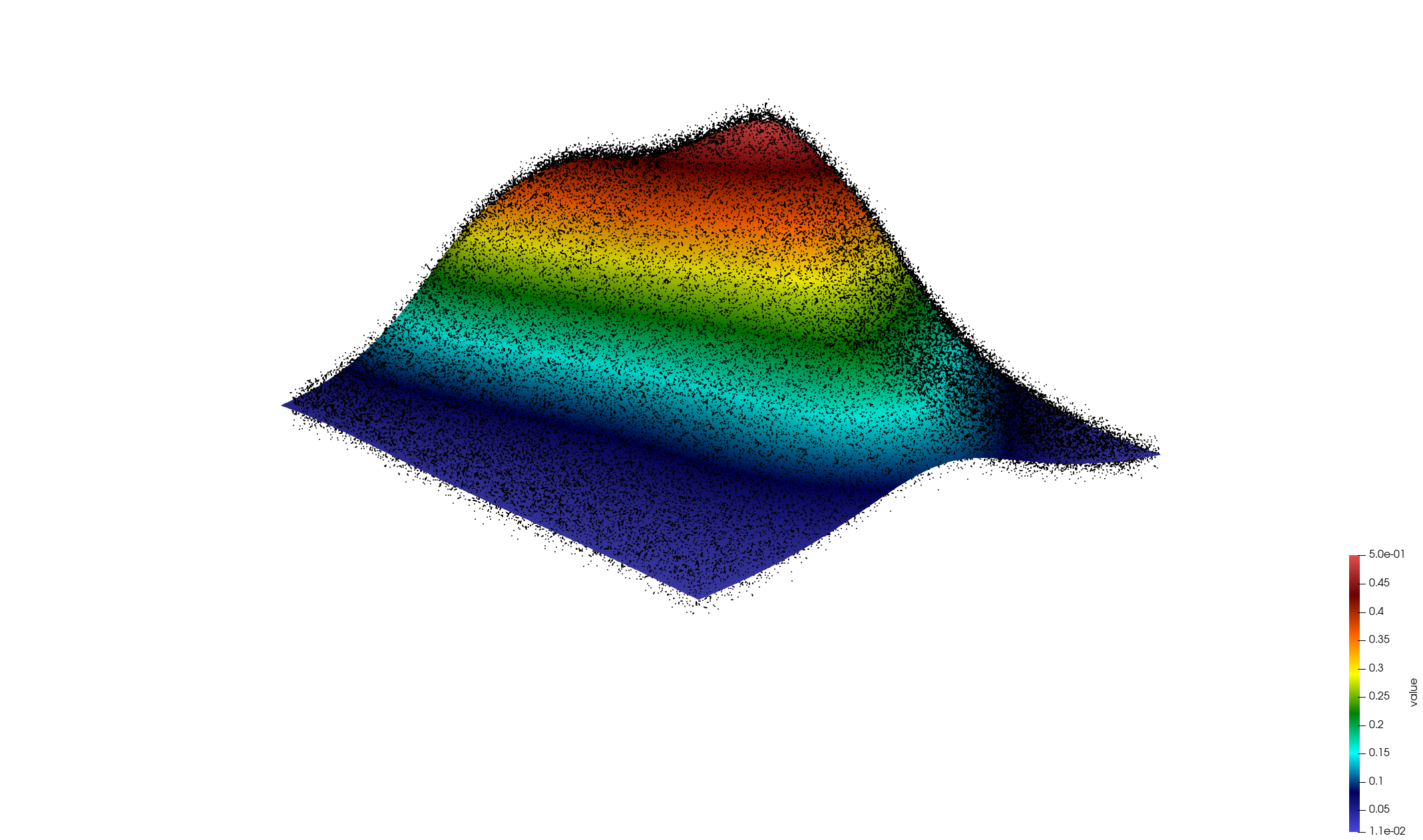}};
\draw(8,6)node{\includegraphics[scale=0.06,clip,trim=230 30 220 100]{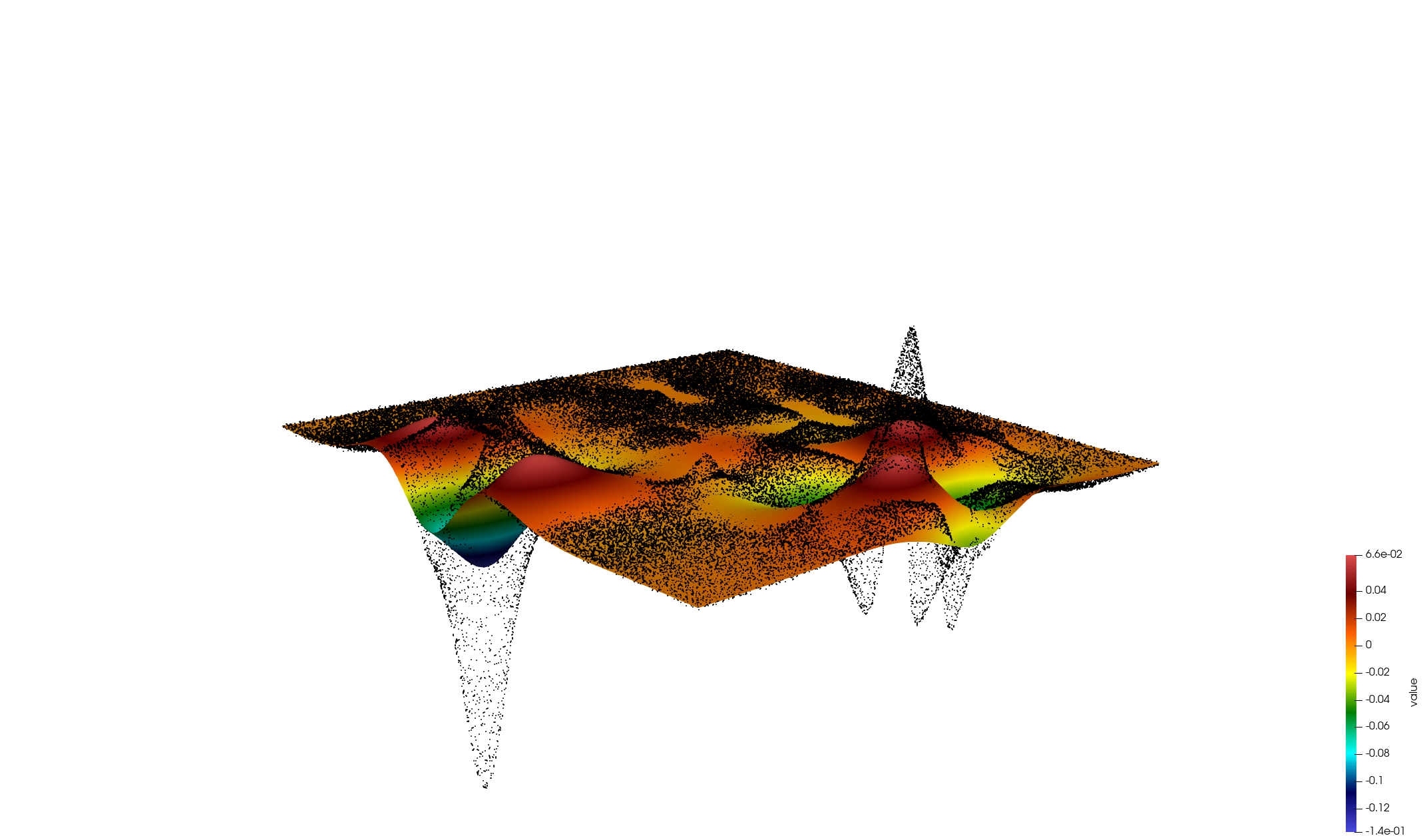}};
\draw(12.3,6)node{\includegraphics[scale=0.06,clip,trim=230 30 220 100]{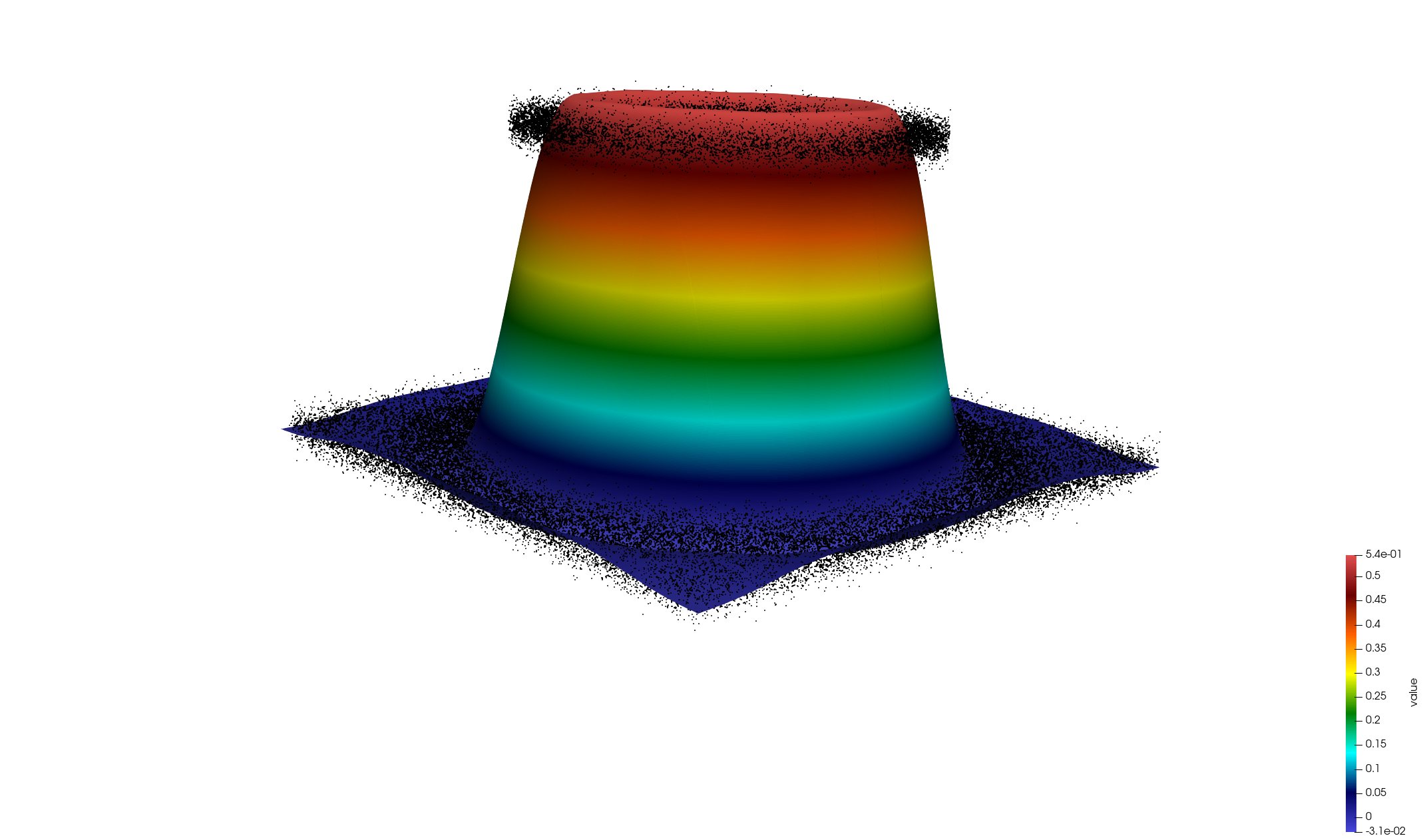}};
\draw(3.7,4)node{\includegraphics[scale=0.06,clip,trim=230 30 220 100]{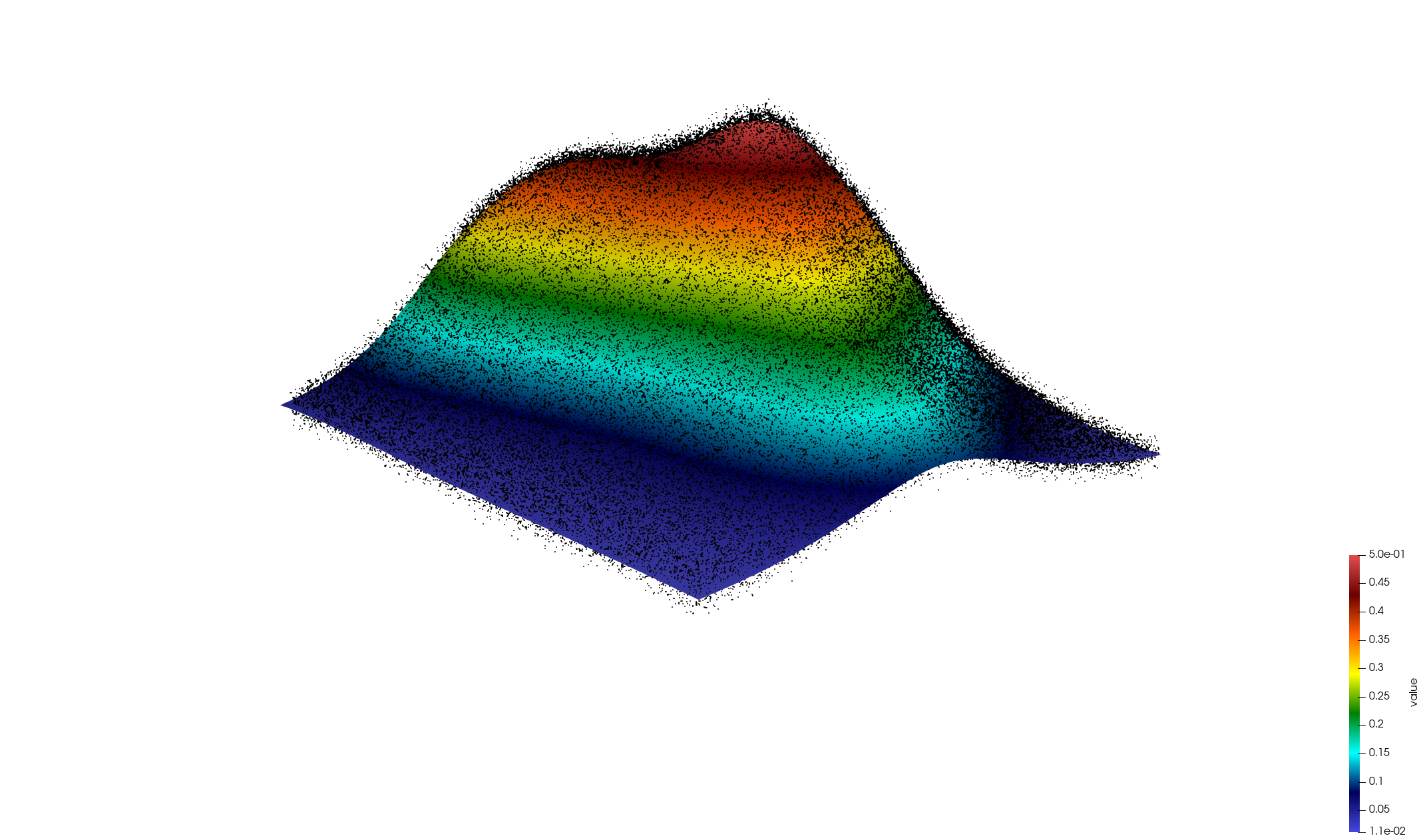}};
\draw(8,4)node{\includegraphics[scale=0.06,clip,trim=230 30 220 100]{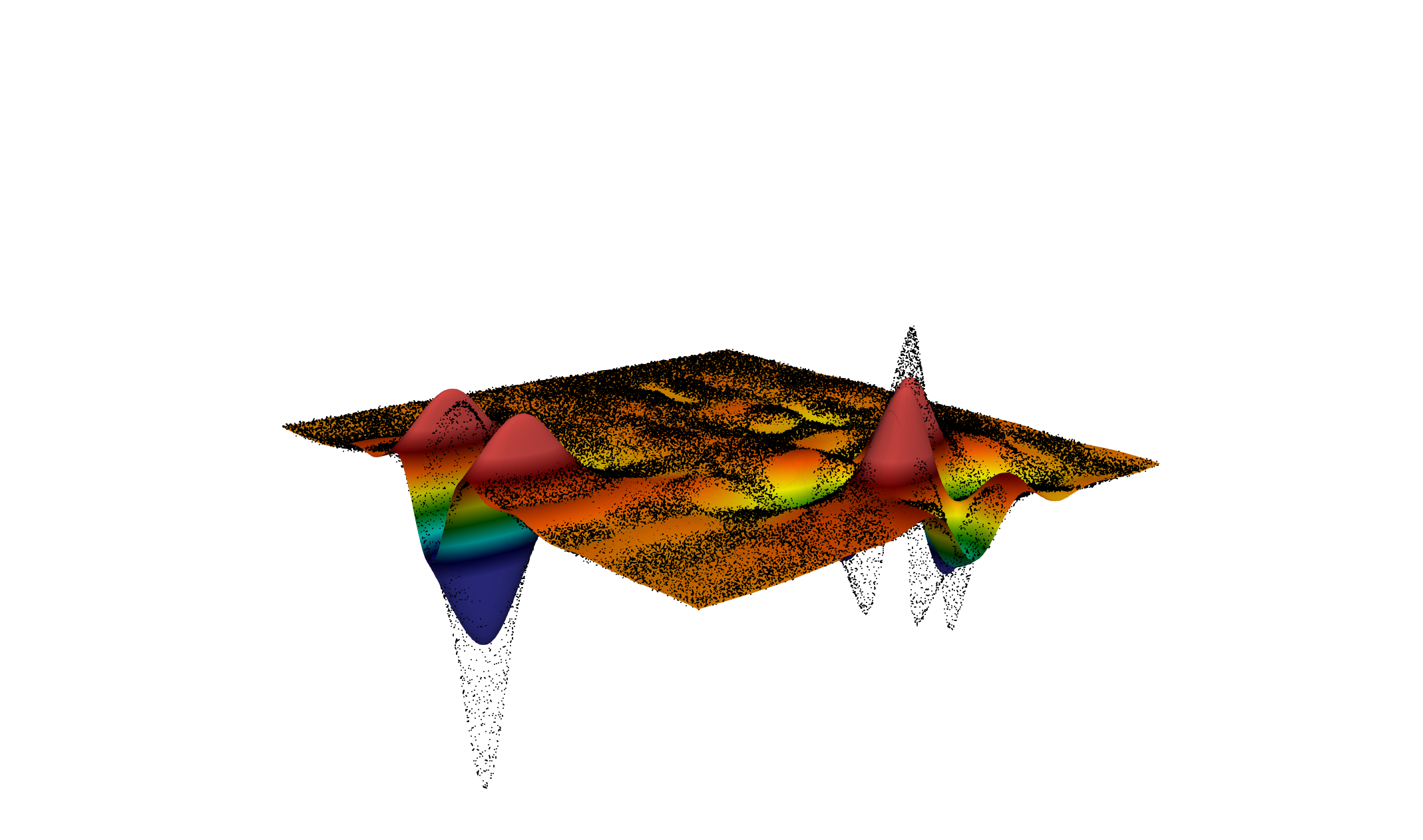}};
\draw(12.3,4)node{\includegraphics[scale=0.06,clip,trim=230 30 220 100]{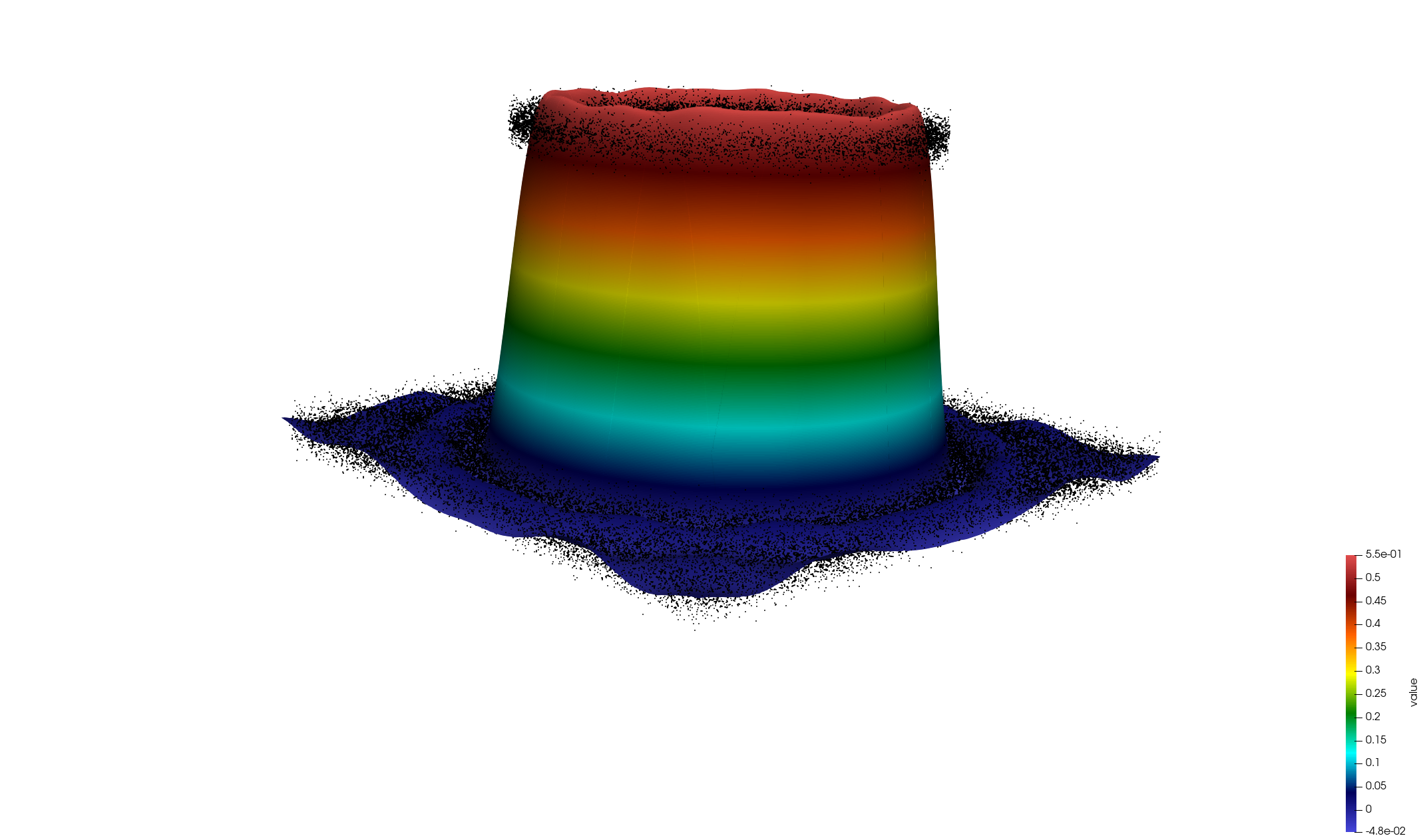}};
\draw(3.7,2)node{\includegraphics[scale=0.06,clip,trim=230 30 220 100]{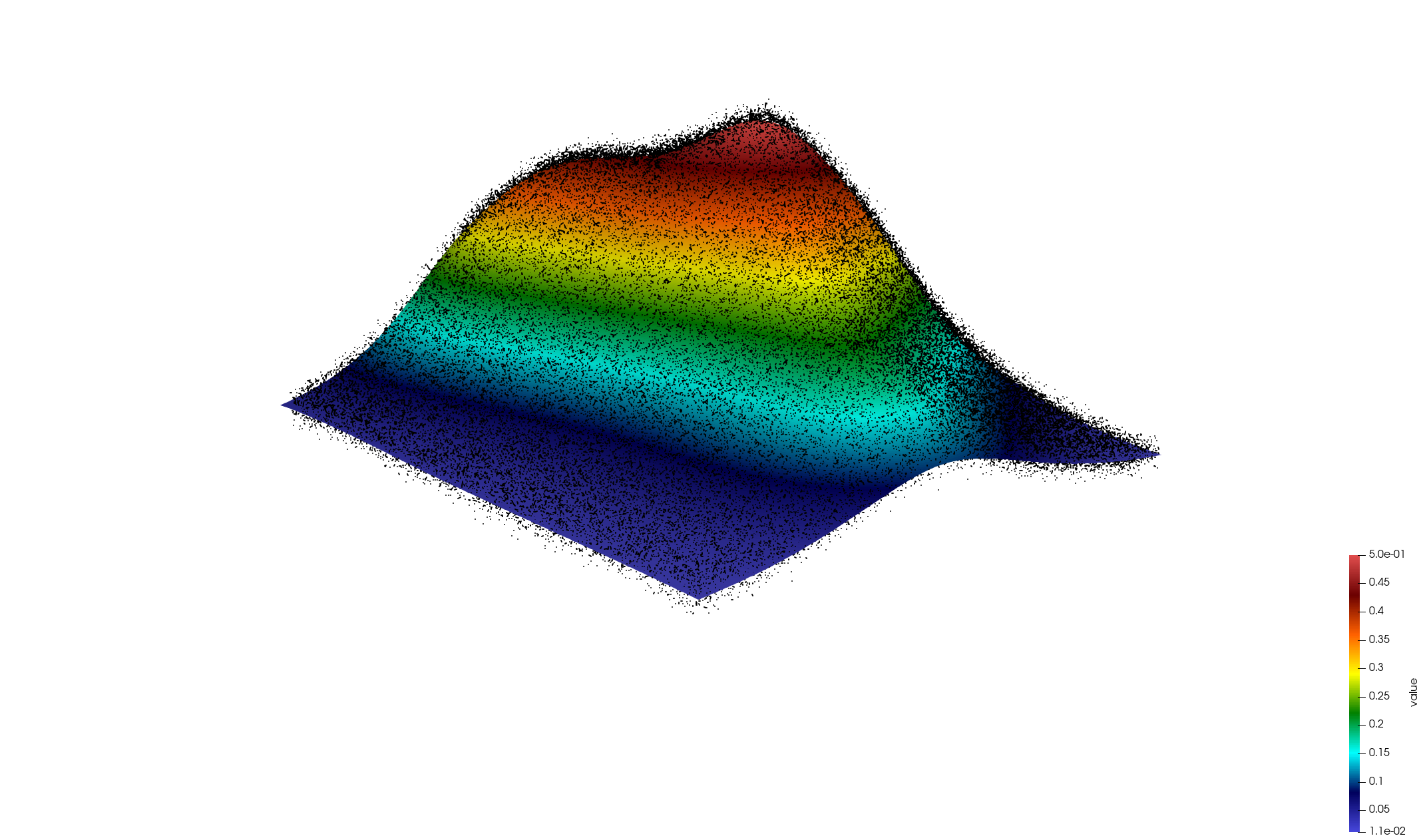}};
\draw(8,2)node{\includegraphics[scale=0.06,clip,trim=230 30 220 100]{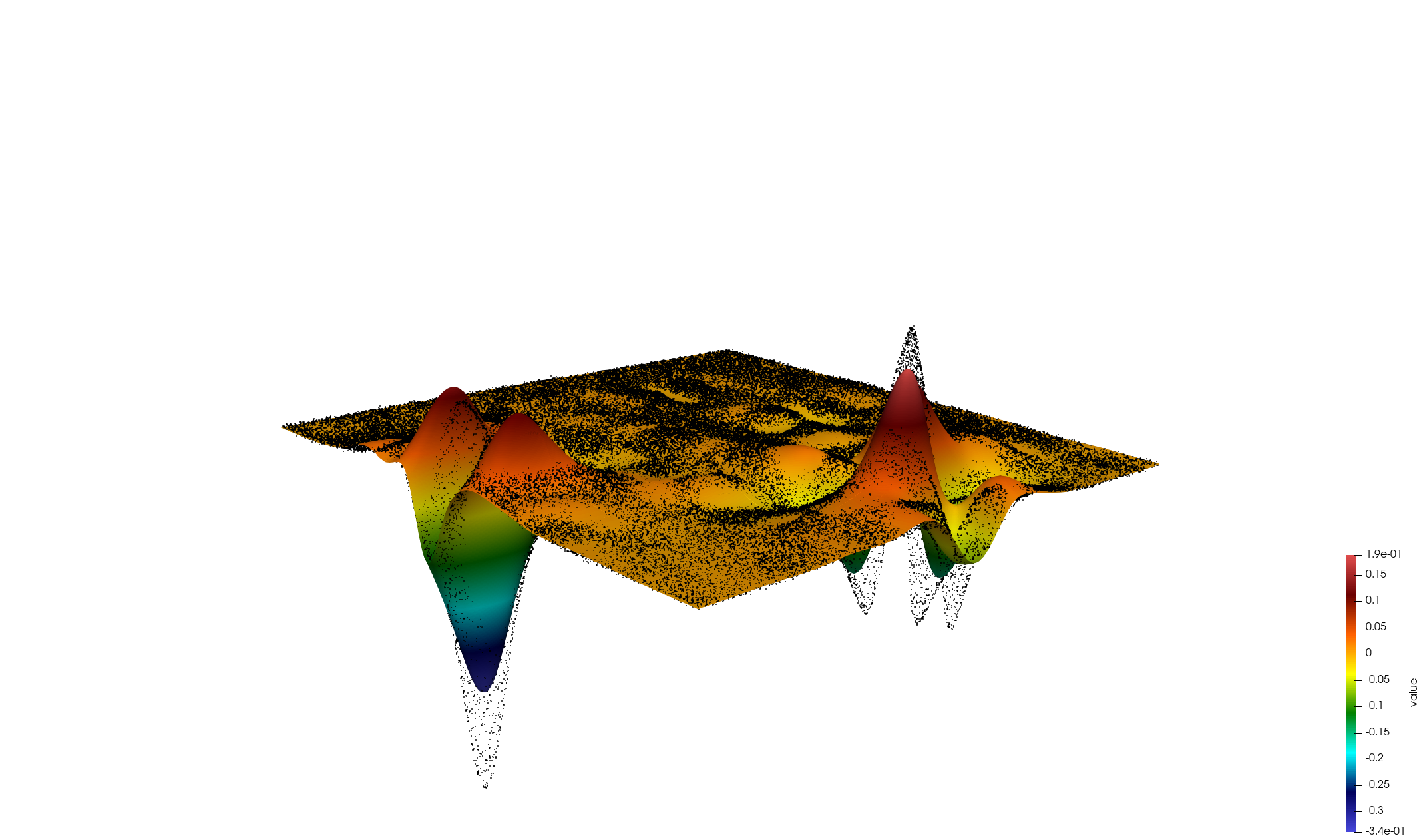}};
\draw(12.3,2)node{\includegraphics[scale=0.06,clip,trim=230 30 220 100]{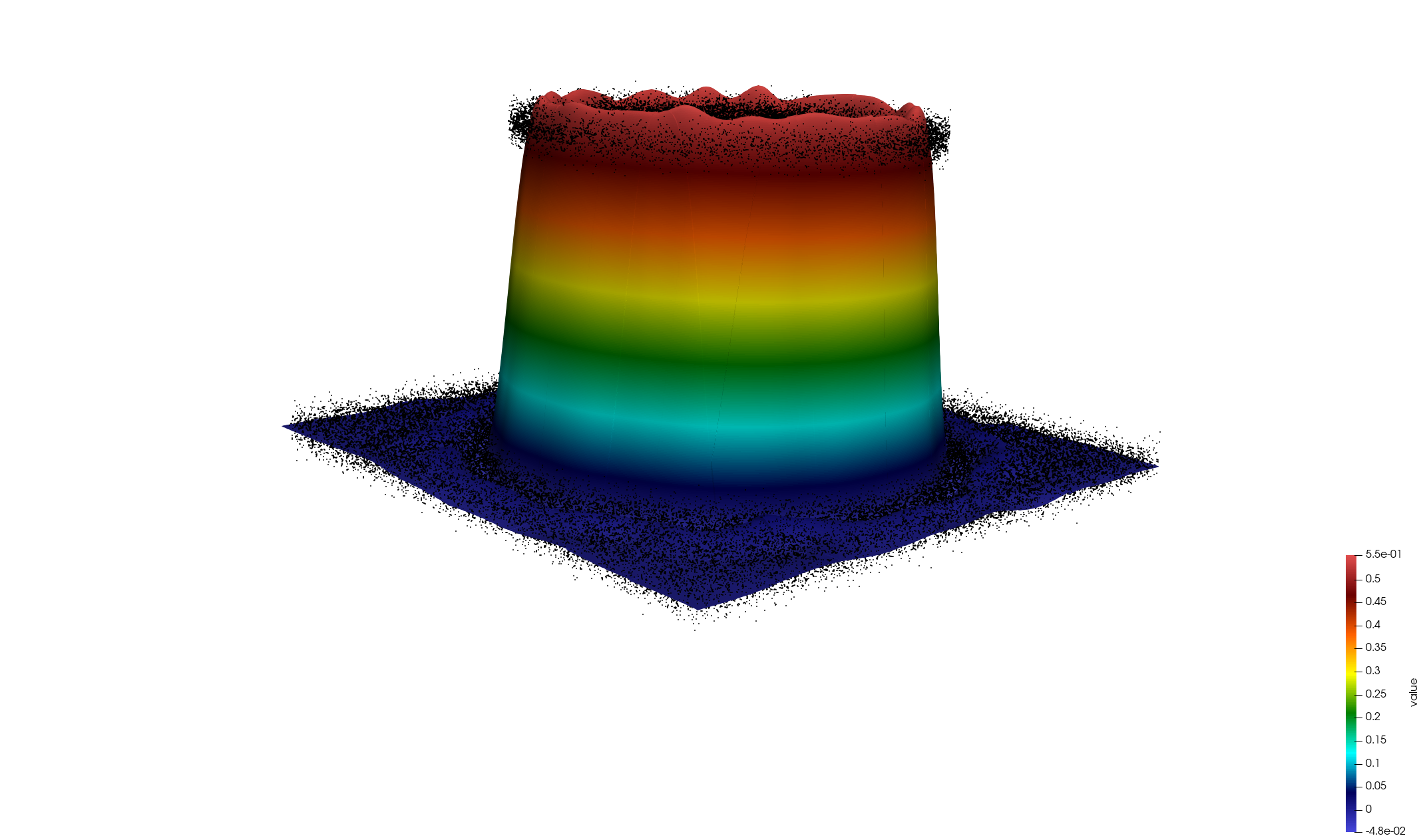}};
\draw(1.4,2)node[rotate=90]{MRSSN};
\draw(1.4,4)node[rotate=90]{MRFISTA};
\draw(1.4,6)node[rotate=90]{FISTA};
\draw(1.4,8)node[rotate=90]{ridge};
\draw(4,9.6)node{$f_{\text{spss}}(X)+{\bs\eta}$};
\draw(8,9.6)node{$f_{\text{spms}}(X)+{\bs\eta}$};
\draw(12.5,9.6)node{$f_{\text{cartoon}}(X)+{\bs\eta}$};
\end{tikzpicture}
\caption{\label{fig:2Ddata}Data samples from the different data generating functions and
reconstructions by the different methods.}
\end{center}
\end{figure*}
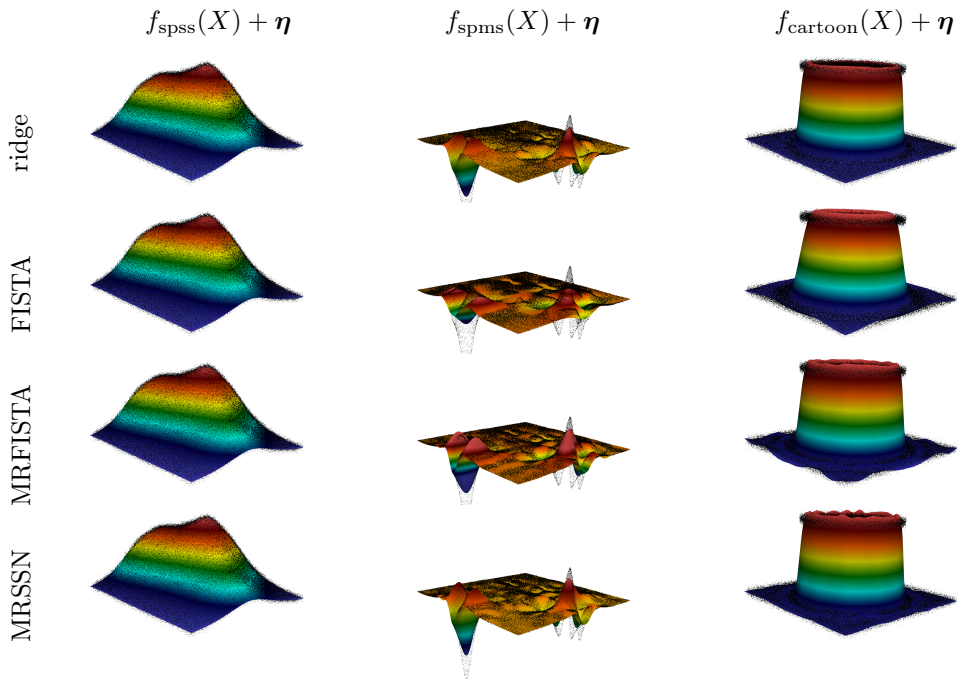
Given a set of uniformly distributed random points 
\[
X=\{{\bs x}_1,\ldots,{\bs x}_N\}\subset[-0.5,0.5]^2,
\]
we consider the Mat{\'e}rn kernel
\[
k_{3/2}(r)=\bigg(1+\frac{\sqrt{3}r}{\ell\sqrt{d}}\bigg)
e^{-\frac{\sqrt{3}r}{\ell\sqrt{d}}}
\]
with correlation length given by \(\ell=0.25\).
To benchmark the algorithms under consideration, we reconstruct
three different data generating functions.

The first one is given by a function that has a sparse representation
in the single-scale basis and is defined as
\[
\begin{aligned}
f_{\text{spss}}({\bs x})\isdef &c_\text{spss}(X)
\sum_{i=1}^{10}k({\bs x}_{m_i},{\bs x}),
\end{aligned}
\]
where \(m_i\in\{1,\ldots, N\}\) are randomly chosen indices.

The second function has a sparse representation in the embedded samplet basis
and reads
\[
\begin{aligned}
f_{\text{spms}}({\bs x})\isdef &c_\text{spms}(X)
\sum_{i=1}^{10}\psi_{m_i}({\bs x}),
\end{aligned}
\]
Herein, \(m_i\in\{10^3,\ldots,10^{4}\}\) are randomly chosen indices
and the embedded samplets \(\psi_{j,k}\) are enumerated in a 
breadth-first-search-like ordering. This means that we randomly select
functions from the low-frequency and middle-frequency part.

In both cases, the normalization constant \(c_{\cdot}(X)\)
is chosen such that the respective functions attain values in 
\([-0.5,0.5]\), that is
\[
  c_\cdot(X)\isdef\frac{1}{2\max_{{\bs x}\in X}f_{\cdot}({\bs x})}.
\]

Finally, we consider the cartoon function
\[
f_{\text{cartoon}}({\bs x})\isdef\begin{cases}0.5,& \|{\bs x}\|_2 < 0.25,\\
0,&\text{else},\end{cases}
\]
as an example that cannot be represented within the basis
of kernel translates
\(\phi_1,\ldots,\phi_N\), see \eqref{eq:HX}.

After generating the function values,
they are perturbed by additive standard gaussian noise
with a noise level of \(5\%\) relative to the euclidean norm of the data.
We consider \(N=10^6\) data points and the regularization parameters are
set to $\lambda/N=2\cdot 10^{-5}$ for the ridge regression and to
$w_i = 2\cdot 10^{-5}$ for \(i=1,\ldots, N\) for the \(\ell^1\)-regularization.
The number of iterations for (MR)FISTA is limited to 10\,000. Otherwise,
the iteration is stopped if the gradient \({\bs d}\) satisfies
\(\|{\bs d}\|_\infty <9\cdot10^{-7}\).
Accordingly, MRSSN is stopped if \(\|{\bs r}\|_\infty <9\cdot10^{-7}\).
We remark that we choose the $\|\cdot\|_\infty$-norm here, since it scales
independently of the number of points $N$. The conjugate gradient method 
for the ridge regression is stopped with a relative
tolerance of \(9\cdot 10^{-7}\).
For all methods, the computation of the samplet compressed kernel matrix
takes $16.04$s and it contains 22 entries per row on average (upper 
triangular part only). The estimated relative  compression error is 
$6.21\cdot 10^{-5}$ in the Frobenius norm.

The obtained interpolants
are evaluated on a grid with \(1\,000\times 1\,000\) points, equalling
\(10^6\) evaluation points in total. The evaluation is performed by 
using the fast multipole method underlying the samplet compression 
with polynomial degree \(4\) and \(\eta=1.2\), see \cite{HM22} for 
the details. It takes about $470$s using 32 
cores in all cases. The reconstructions 
for the different methods under consideration are depicted in 
Figure~\ref{fig:2Ddata}. The columns correspond to the different
data generating functions, while the rows correspond to the different 
methods. A subsample of \(10^5\) data points is indicated by the 
black dots. 

In all examples, the maximum number of iterations of 
(MR)FISTA was reached before the convergence criterion 
was met, compare Table~\ref{table:SS}, Table~\ref{table:SM},
and Table~\ref{table:CF}.
\begin{table}[!t]
\centering
\begin{tabular}{|l|l|l|l|l|}
\hline
 & ridge & FISTA & MRFISTA & MRSSN\\\hline
iterations & $103$& $10\,000$& $10\,000$ & $411$\\\hline
comp. time& $4.51$s & $1\,777.42$s & $501.23$s& $58.48$s\\\hline
final $|\Acal|$ & $973\,731$ & $32\,963$& $111$ & $101$\\\hline
error & $8.3\cdot10^{-8}$
& $1.2\cdot10^{-2}$
& $1.2\cdot10^{-2}$
& $2.0\cdot10^{-9}$
\\\hline
\end{tabular}
\vspace*{0.25em}
\caption{\label{table:SS}Results for sparse single-scale data.}
\centering
\begin{tabular}{|l|l|l|l|l|}
\hline
 & ridge & FISTA & MRFISTA & MRSSN\\\hline
iterations & $170$& $10\,000$& $10\,000$ & $422$\\\hline
comp. time& $11.3$s & $1\,857.19$s & $506.86$s & $57.86$\\\hline
final $|\Acal|$ & $827\,864$ & $35\,194$& $491$ & $163$\\\hline 
error &$1.2\cdot10^{-8}$
& $2.6\cdot10^{-2}$
& $1.6\cdot10^{-2}$
& $2.3\cdot10^{-7}$
\\\hline
\end{tabular}
\vspace*{0.25em}
\caption{\label{table:SM}Results for sparse multiresolution data.}
\centering
\begin{tabular}{|l|l|l|l|l|}
\hline
                  & ridge & FISTA & MRFISTA & MRSSN\\\hline
iterations        & $118$& $10\,000$& $10\,000$ & $646$\\\hline
comp. time&       $5.17$s & $1970.76$s & $607.05$s& $80.46$s\\\hline
final $|\Acal|$ & $978\,740$ & $218\,570$& $921$ & $261$\\\hline
error             &$9.6\cdot10^{-8}$ & $7.1\cdot10^{-2}$ & $5.6\cdot10^{-2}$ & $3.4\cdot10^{-7}$
\\\hline
\end{tabular}
\vspace*{0.25em}
\caption{\label{table:CF}Results for cartoon function data.}
\end{table}
The tables also show the cardinality of the final active index 
sets \(\Acal\). The number for the ridge regularization is obtained 
counting the coefficients whose modulus is larger than the 
threshold as for the \(\ell^1\)-regularization. It can be seen 
that MRSSN leads throughout to the sparsest solution, 
while the computation time is rather moderate compared 
to (MR)FISTA.

A qualitative comparison of the different methods in terms of
their reconstruction capabilities can be found in the visualizations
in Figure~\ref{fig:2Ddata}. Except for the sparse single-scale 
example, FISTA performs significantly worse then the 
multiresolution competitors. For the sparse multiresolution example, 
MRSSN performs best, while the ridge regression is slightly worse.
For the cartoon function, the ridge regularization provides the best tradeoff between
a smooth solution and a solution which fits the data, while the results from
MRFISTA and MRSSN show some oscillations at the jump.

\begin{figure*}[!t]
\begin{center}
\begin{tikzpicture}
\draw(4,7)node{\includegraphics[scale=0.07,clip,trim=440 60 440 60]{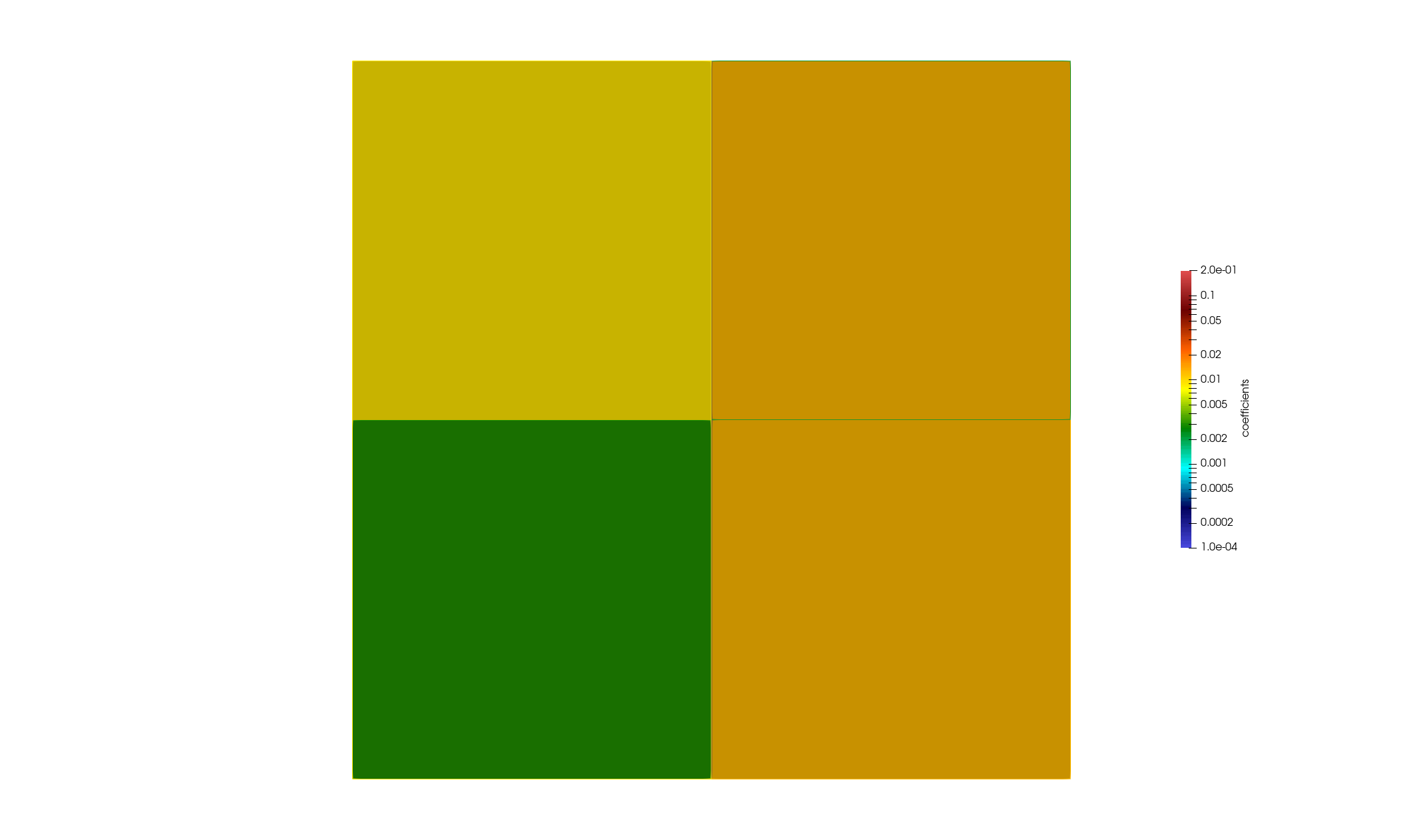}};
\draw(8,7)node{\includegraphics[scale=0.07,clip,trim=440 60 440 60]{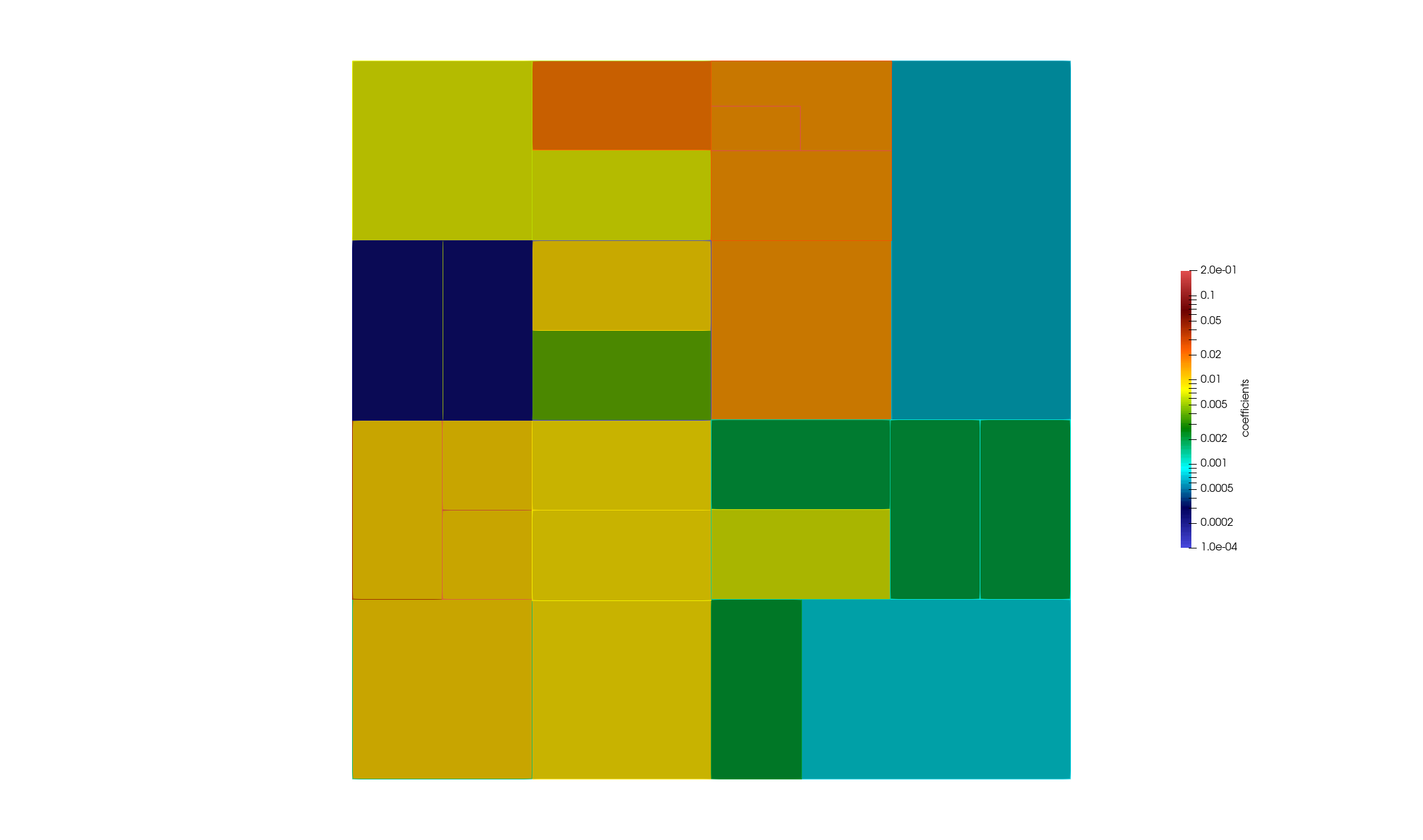}};
\draw(12,7)node{\includegraphics[scale=0.07,clip,trim=440 60 440 60]{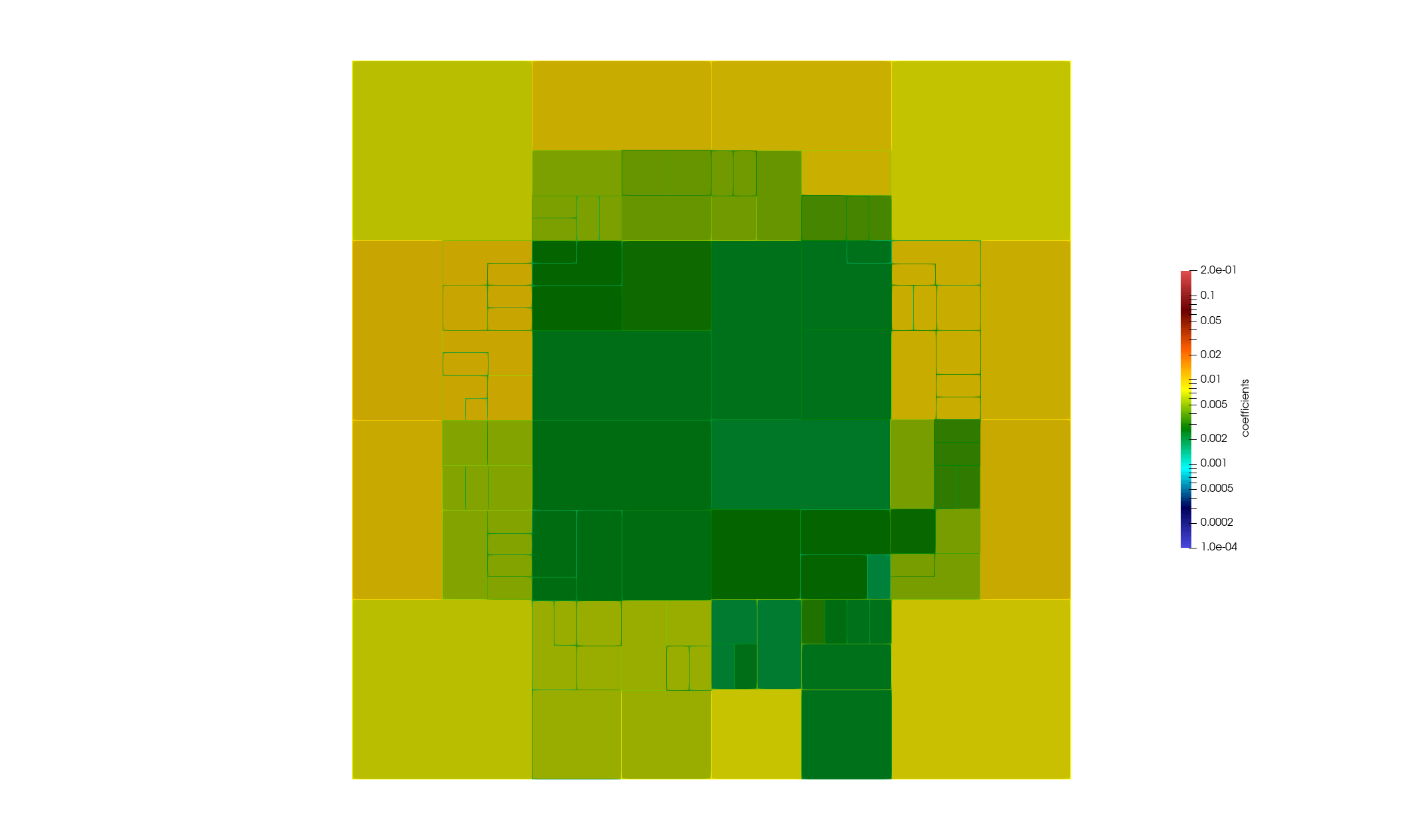}};
\draw(4,4)node{\includegraphics[scale=0.07,clip,trim=440 60 440 60]{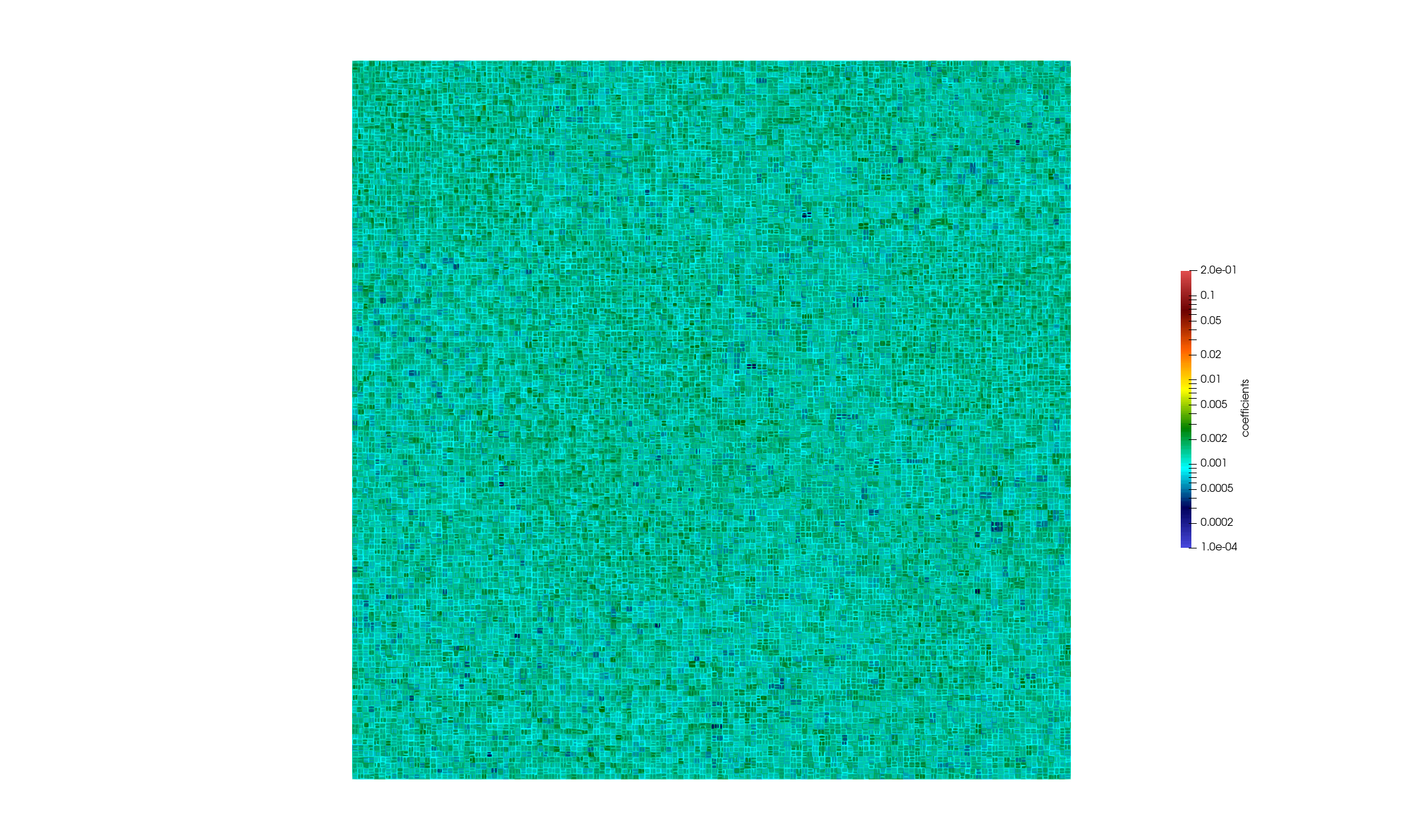}};
\draw(8,4)node{\includegraphics[scale=0.07,clip,trim=440 60 440 60]{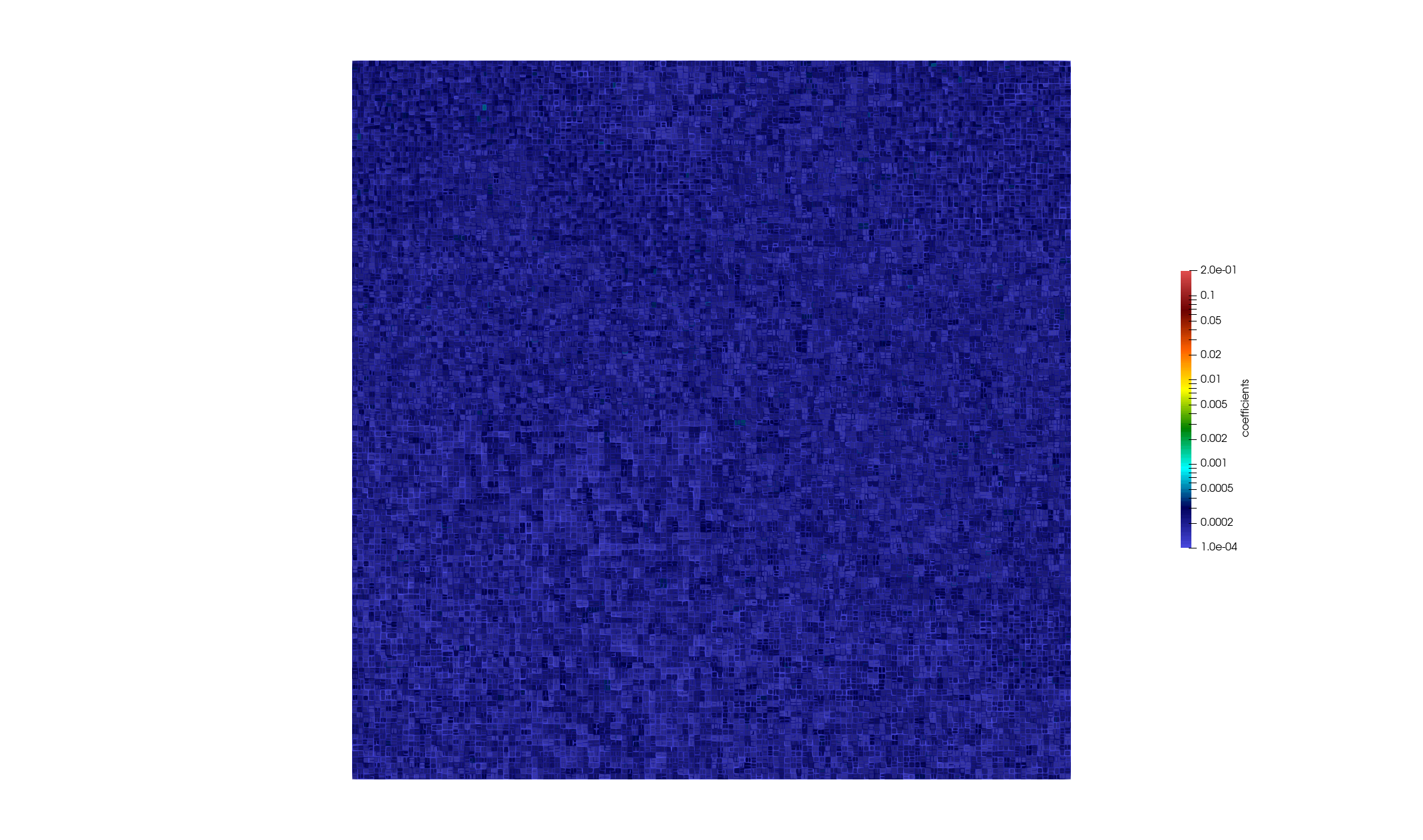}};
\draw(12,4)node{\includegraphics[scale=0.07,clip,trim=440 60 440 60]{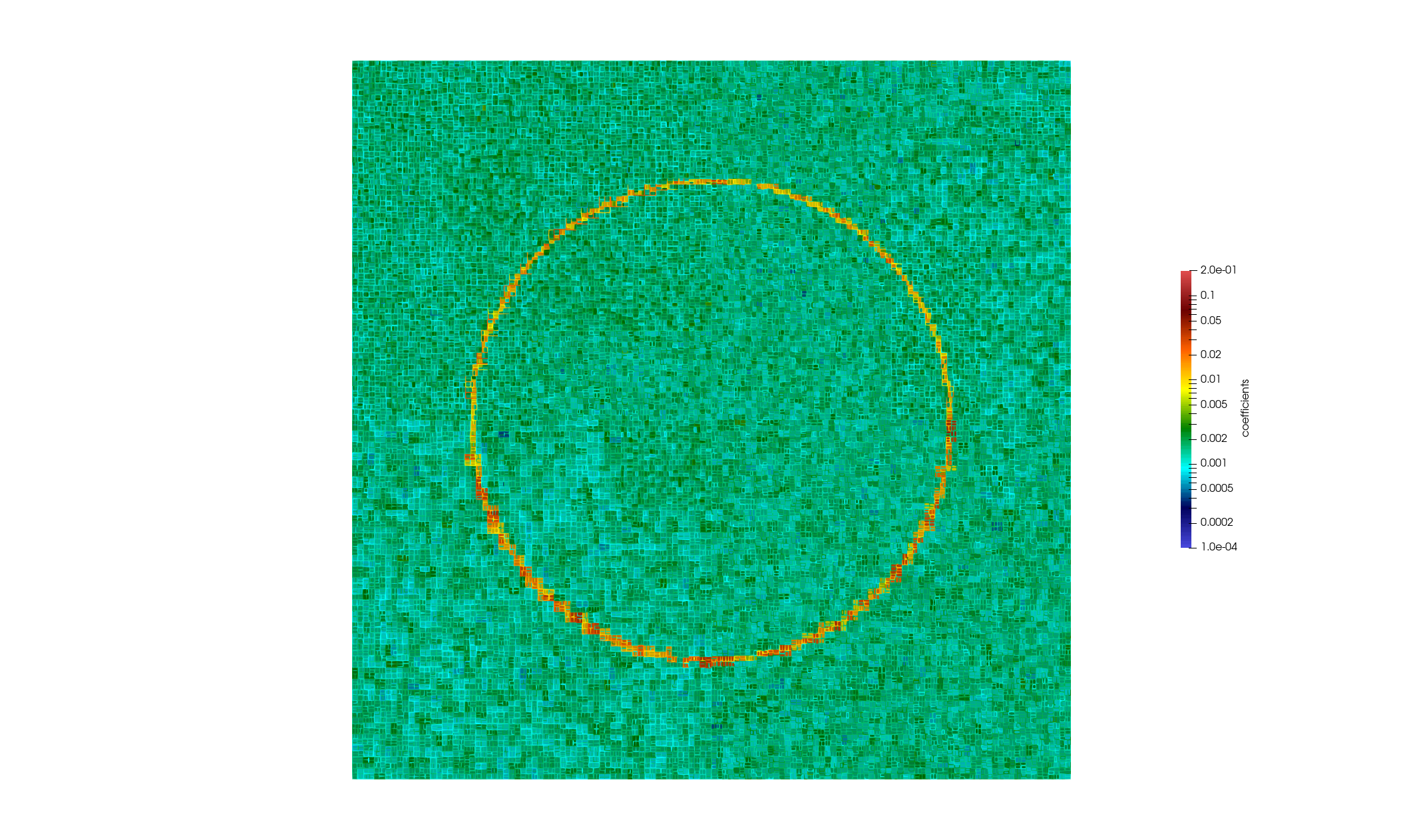}};
\draw(4,1)node{\includegraphics[scale=0.07,clip,trim=440 60 440 60]{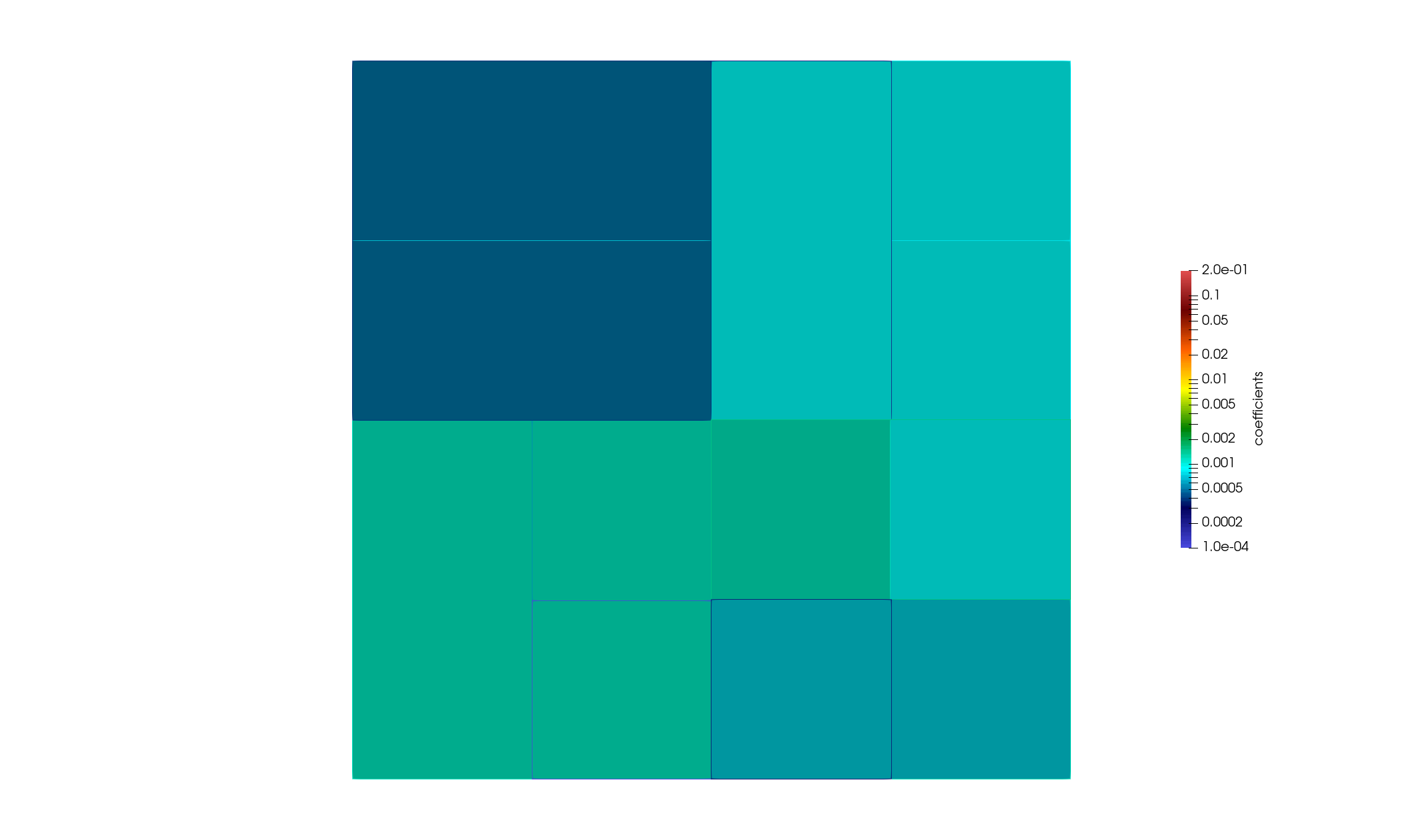}};
\draw(8,1)node{\includegraphics[scale=0.07,clip,trim=440 60 440 60]{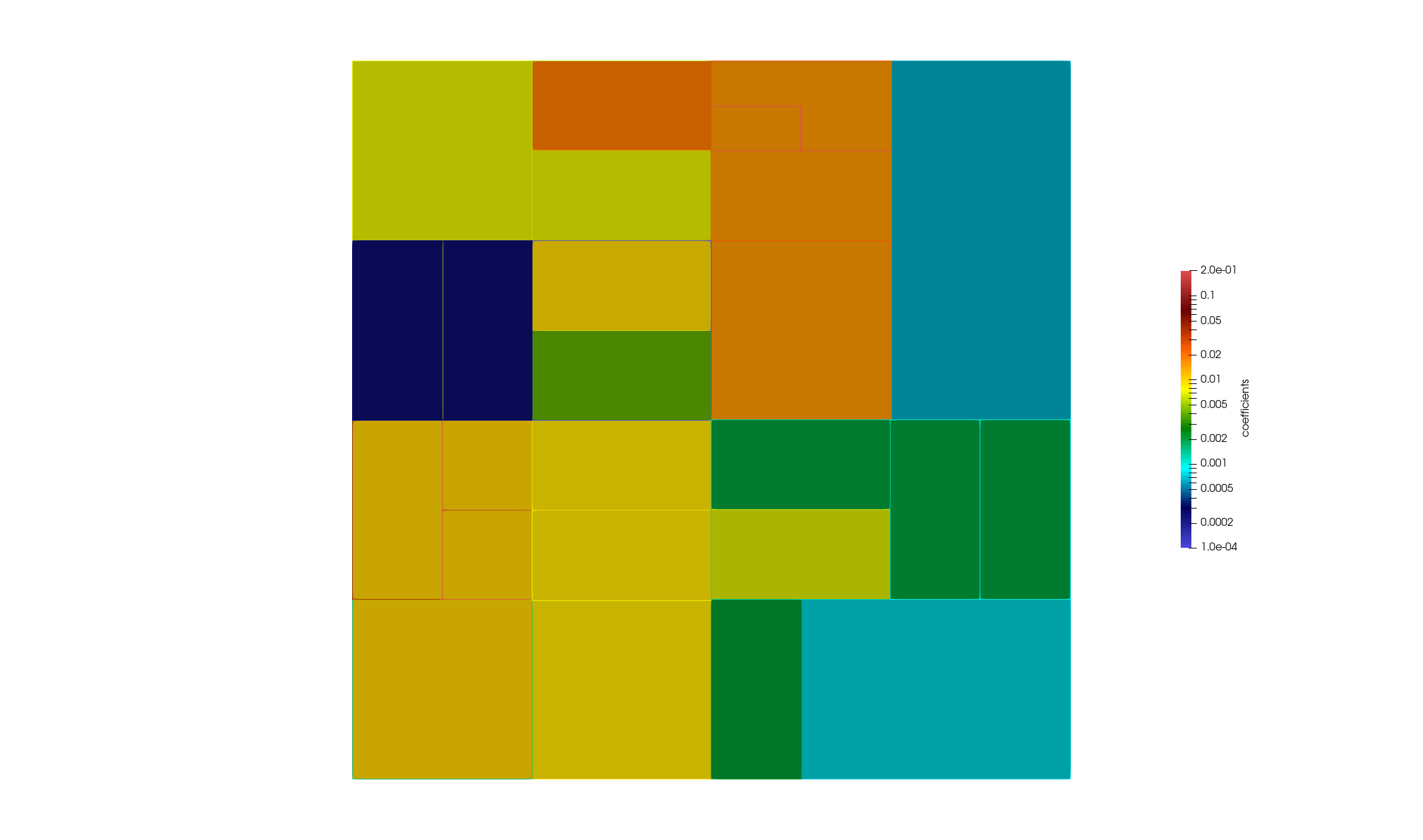}};
\draw(12,1)node{\includegraphics[scale=0.07,clip,trim=440 60 440 60]{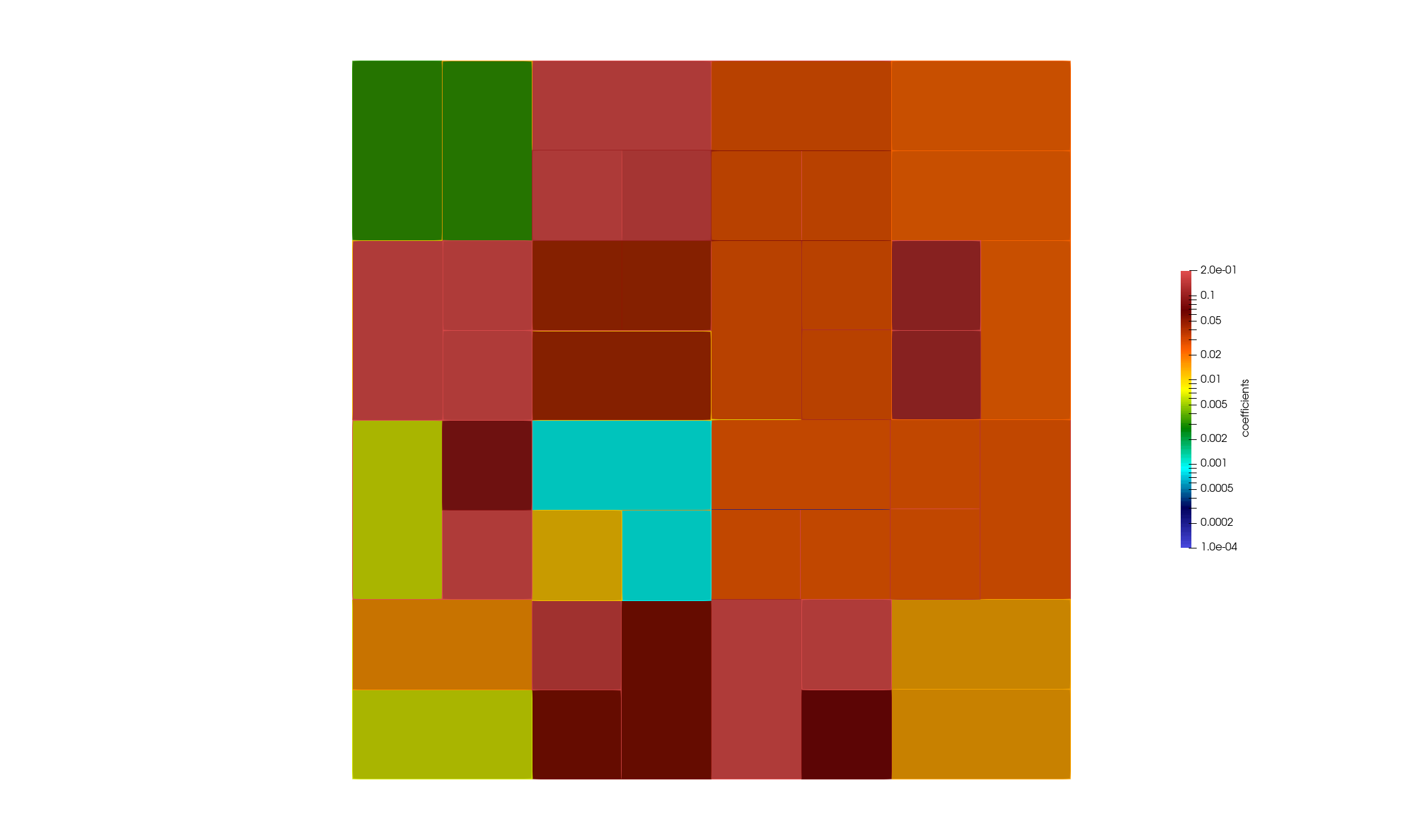}};
\draw(2,1)node[rotate=90]{SSN};
\draw(2,4)node[rotate=90]{ridge};
\draw(2,7)node[rotate=90]{data};
\draw(4,8.8)node{$f_{\text{spss}}(X)+{\bs\eta}$};
\draw(8,8.8)node{$f_{\text{spms}}(X)+{\bs\eta}$};
\draw(12,8.8)node{$f_{\text{cartoon}}(X)+{\bs\eta}$};
\draw(15,4)node{\includegraphics[rotate=0,scale=0.2,clip,trim=1770 420 230 380]{Images/l1_2D_TSSNcoeffs_rhs1.png}};
\end{tikzpicture}
\caption{\label{fig:2DdataCoeff}Sparsity pattern of the data (coefficients that
are larger than 1\% of the maximum coefficient are shown) and sparsity patterns 
of the corresponding solutions
visualized by the supports
of the active samplets
in case of the ridge regression and the SSN.}
\end{center}
\end{figure*}

Finally, we provide a comparison between the coefficients obtained by
hard-thresholding the ridge regularized solution and the solution from MRSSN in
Figure~\ref{fig:2DdataCoeff}. The first row shows the dominant coefficients
of the signal indicated by their supports. The second row shows the coefficients
obtained from the reconstruction based on ridge regularization, while the last row
shows the sparse reconstruction using MRSSN. Here it can particularly be seen 
that the better fit of the ridge regularization for the cartoon function can be explained
by the very high resolution in terms of samplet coefficients at the interface.

\subsection{Surface reconstruction}\label{ss:Surf}
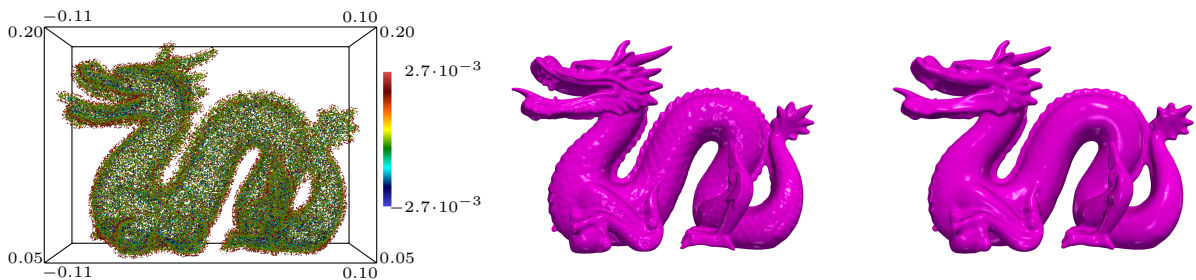
\begin{figure*}[!t]
\begin{center}
\begin{tikzpicture}
\draw(15.8,0)node{\includegraphics[scale=0.105,clip,trim=150 80 200 120]{
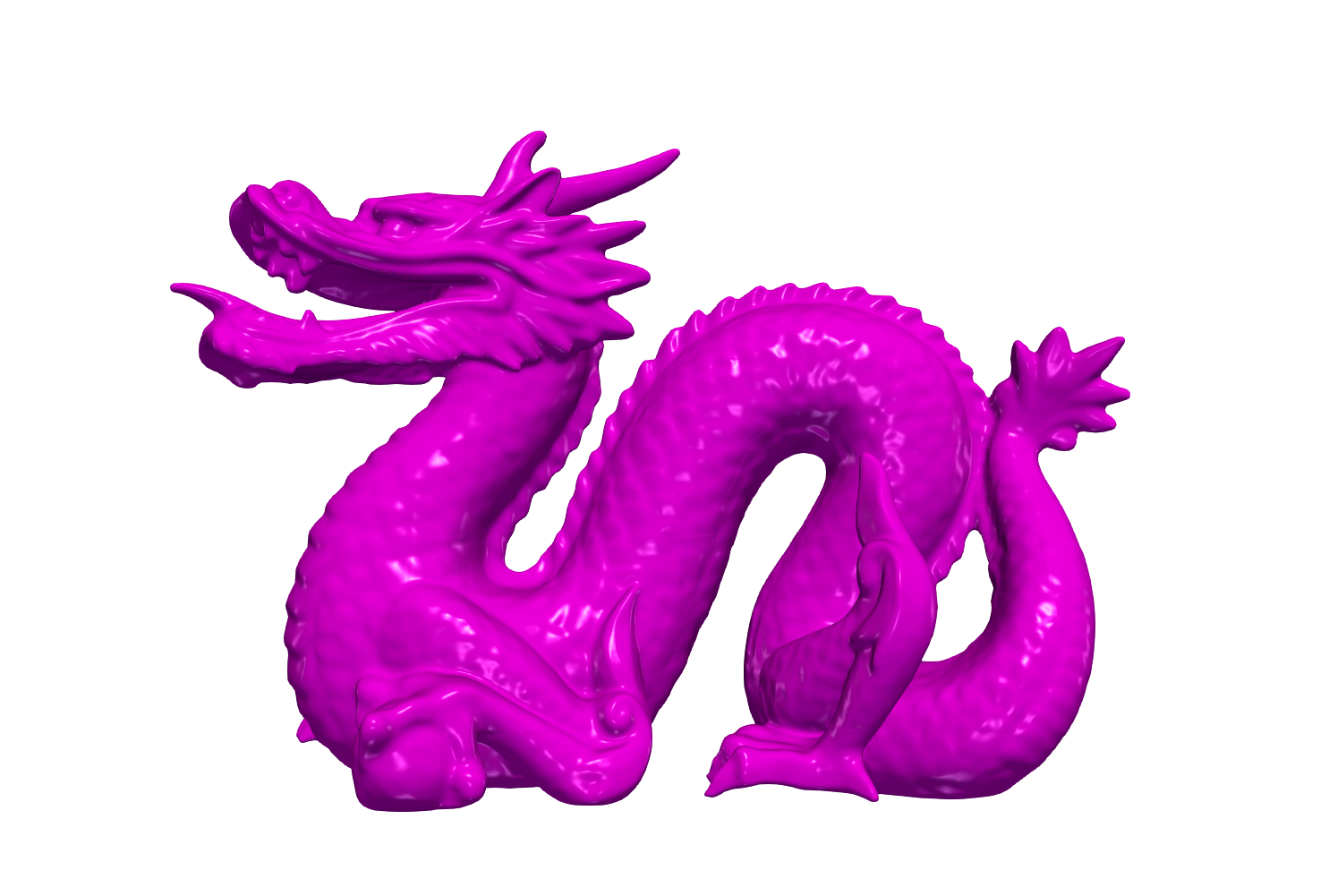}};
\draw(20.8,0)node{\includegraphics[scale=0.105,clip,trim=150 80 200 120]{
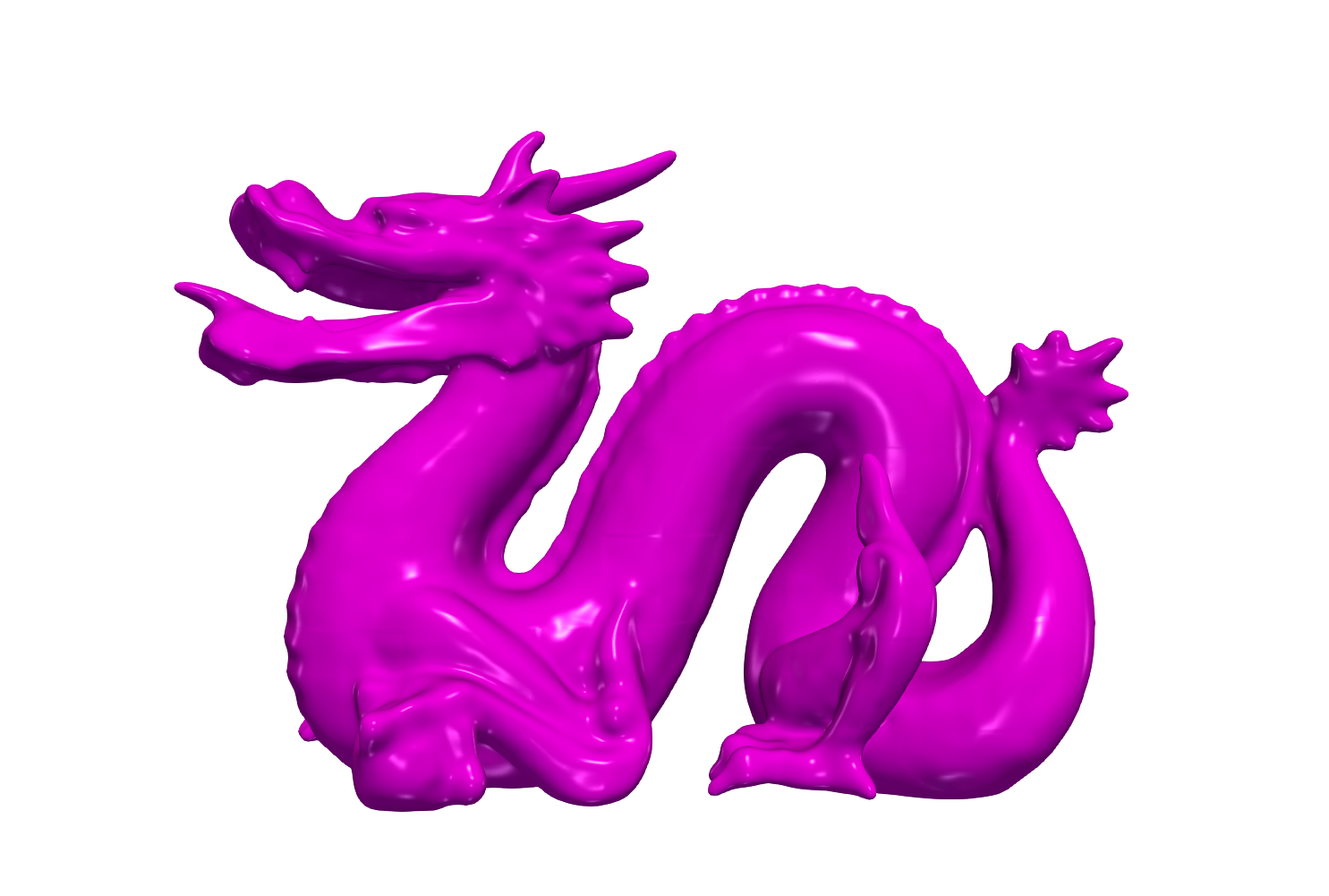}};
\draw(10,0)node{\includegraphics[scale=0.1,clip,trim=375 315 310 315]{
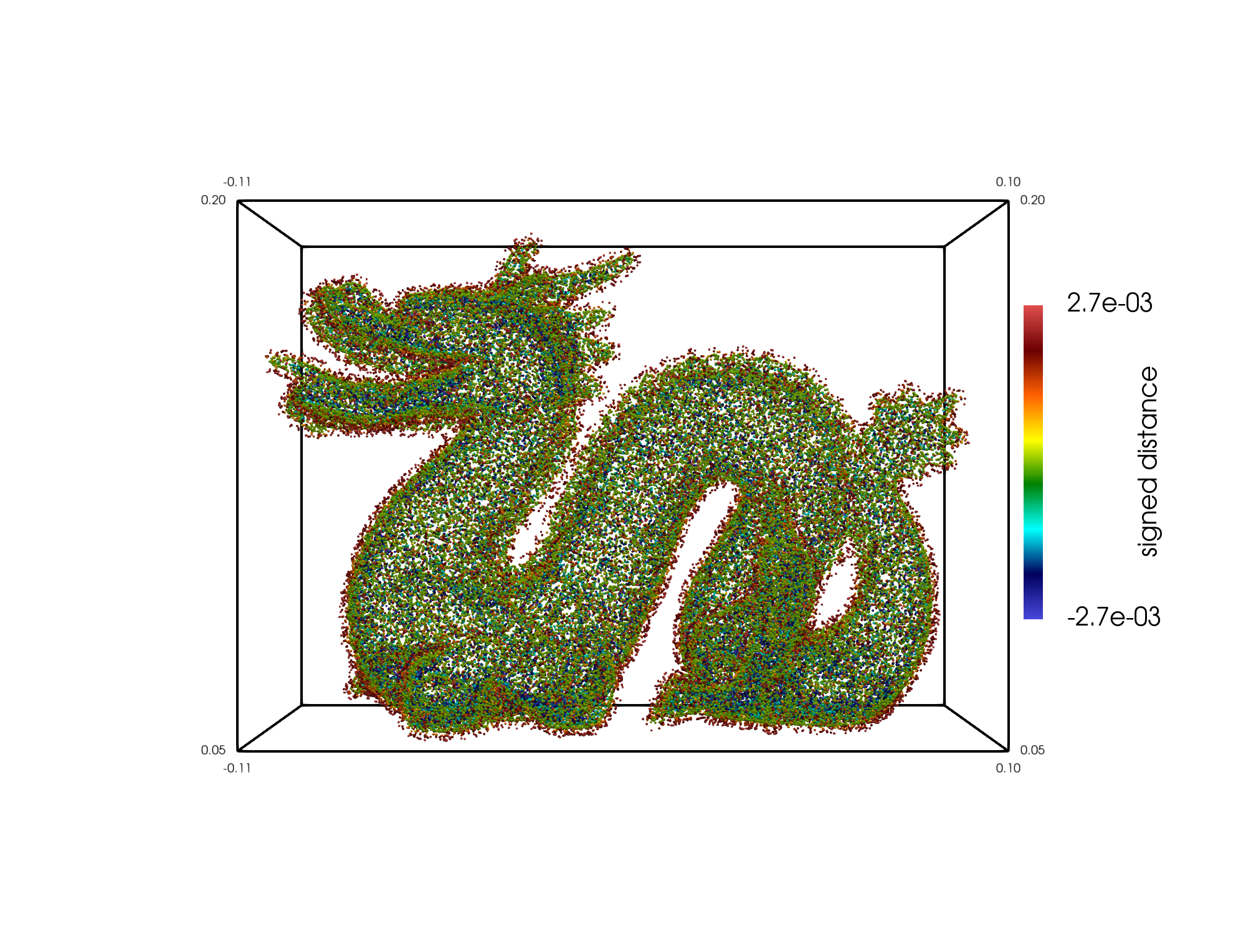}};
\draw(7.45,-1.5)node{\tiny$0.05$};
\draw(8,-1.7)node{\tiny$-0.11$};
\draw(12.34,-1.5)node[fill=white,inner sep=0.5,outer sep=0]{\tiny$0.05$};
\draw(11.85,-1.7)node{\tiny$0.10$};
\draw(12.34,1.5)node[fill=white,inner sep=0.5,outer sep=0]{\tiny$0.20$};
\draw(11.85,1.7)node{\tiny$0.10$};
\draw(7.45,1.5)node{\tiny$0.20$};
\draw(8,1.7)node{\tiny$-0.11$};
\draw(12.85,1)node{\tiny$\phantom{-}2.7\!\cdot\! 10^{-3}$};
\draw(12.85,-0.8)node{\tiny$-2.7\!\cdot\! 10^{-3}$};
\end{tikzpicture}
\caption{\label{fig:StanfordDragon}Surface reconstruction from measurements
of the signed distance function. The left panel shows 
a subsample of the used data, the reconstruction by
ridge regression is found in the middle, and the reconstruction by $\ell^1$-regularization
is in the right panel.}
\end{center}
\end{figure*}
\begin{figure*}[!t]
\begin{center}
\begin{tikzpicture}
\draw(15.8,0)node{\includegraphics[scale=0.1,clip,trim=400 70 400 70]{
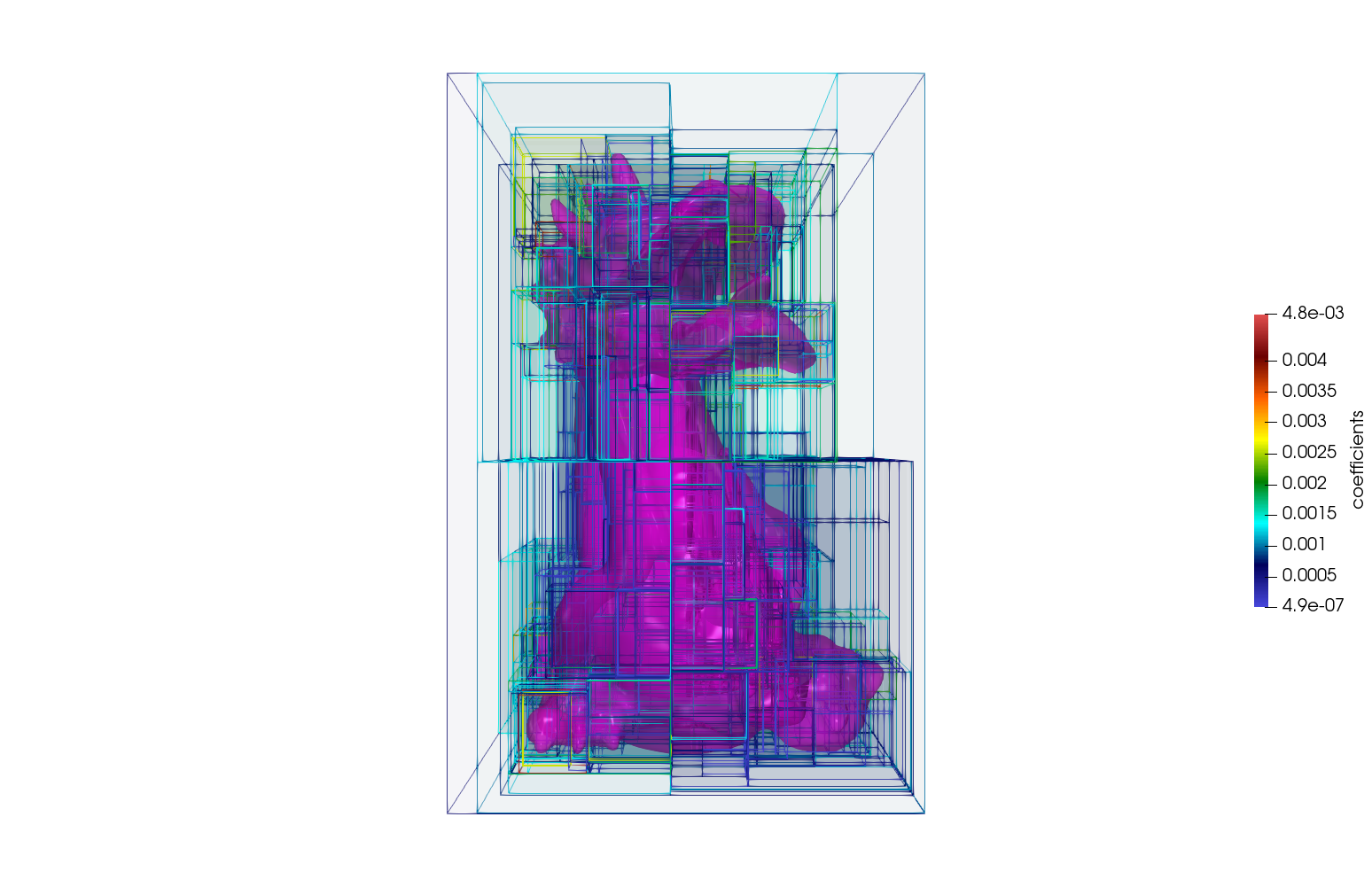}};
\draw(19.8,0)node{\includegraphics[scale=0.094,clip,trim=120 50 145 50]{
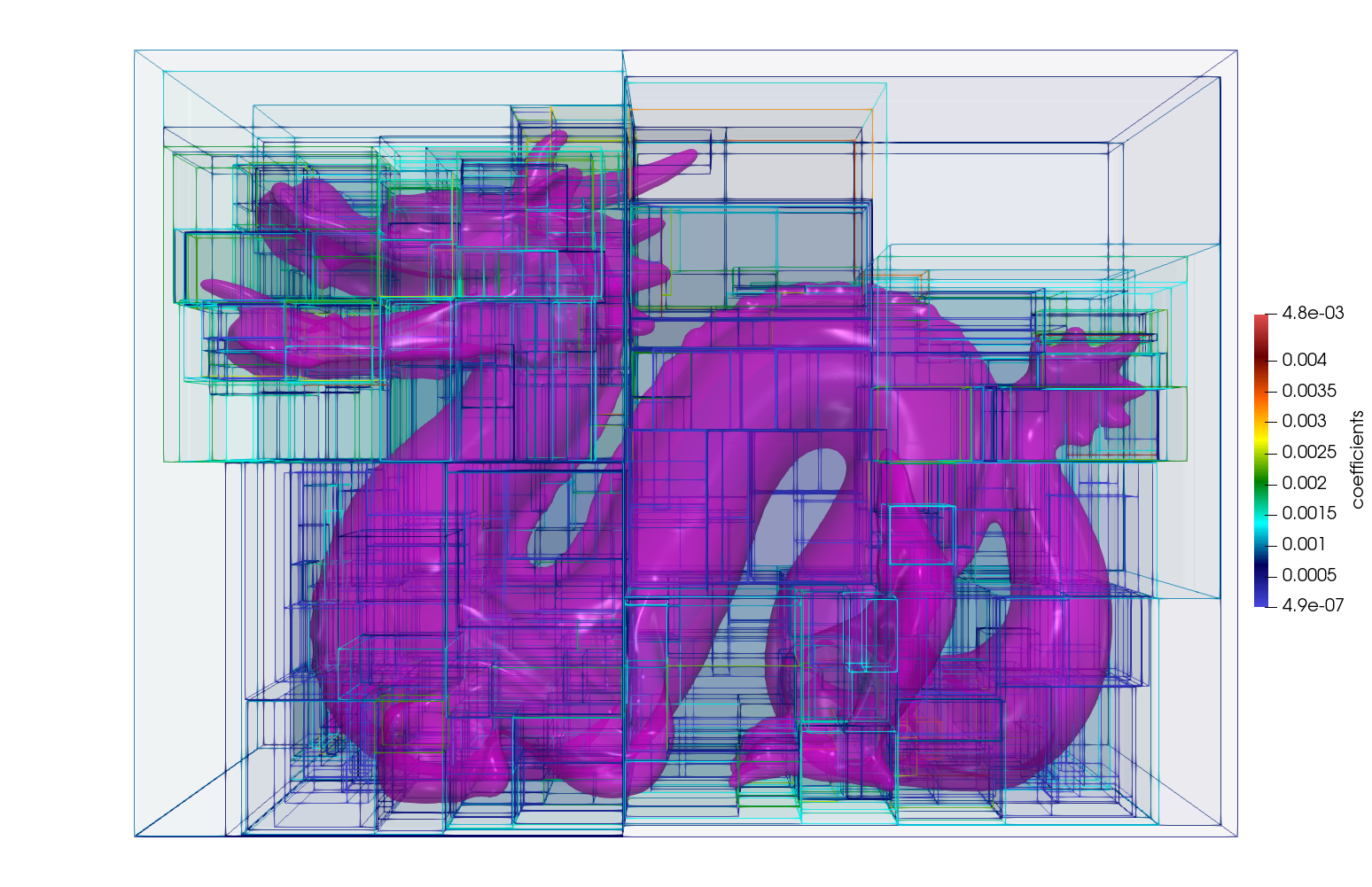}};
\draw(23,0)node{\includegraphics[scale=0.2,clip,trim=1400 300 0 300]{
./Images/l1surf_coeff.png}};
\end{tikzpicture}
\caption{\label{fig:StanfordDragonCoeff}Clusters containing the active coefficients
for the case of the $\ell^1$-regularization.}
\end{center}
\end{figure*}
In this example, we consider the reconstruction of a surface given by samples
of a signed distance function, as examined earlier in \cite{CarrEtAl01}.
We replace the greedy subsampling procedure from the aforementioned reference
by imposing sparsity constraints on the coefficients of the reconstruction.
As data set, we consider $5\cdot{10^5}$ randomly selected points at the
surface of the Stanford dragon and $5\cdot{10^5}$ randomly selected points
at a distance in [0.002,0.01] relative to the diameter of the Stanford dragon's
bounding box, compare the left panel of Figure~\ref{fig:StanfordDragon}, where
also a subsample of $10^5$ data points is shown. 
The total number of points is hence again \(N=10^6\).

For the surface reconstruction, we use the exponential
kernel
\[
k_{\exp}(r)=e^{-\frac{r}{\ell\sqrt{d}}}
\]
with correlation length \(\ell=0.03\). The computation of the 
samplet compressed kernel matrix takes $313$s
and the estimated relative compression error in the Frobenius 
norm is $2.62\cdot 10^{-5}$. The compressed matrix contains 
$1.51\cdot10^3$ entries per row on average (upper triangular
part only). The regularization parameter is set to \(\lambda/N
= 5\cdot 10^{-9}\) for the ridge regression and \(w_i=5\cdot 10^{-9}\)
for \(i=1,\ldots,N\) for the \(\ell^1\)-regularization. The benchmark 
results in Subsection~\ref{subs:bench} suggest that MRSSN 
performs best. Therefore, we will exclusively rely on this 
algorithm in what follows. The conjugate gradient method 
for the ridge regression is stopped with a tolerance of 
\(9\cdot 10^{-7}\) while the MRSSN is stopped with 
\(\|{\bs r}\|_\infty<9\cdot10^{-7}\). The latter yields 
a sparse solution with only \(\|{\bs\beta}\|_0 =6\,233\) 
non-vanishing coefficients. The conjugate gradient method 
takes $1.33\cdot 10^4$s to converge, while the MRSSN 
takes about four times as long with $2.02\cdot 10^4$s (single 
threaded times with parallel assembly of \({\bs M}_{\Acal\Acal}\)). 

The solution is evaluated on a structured grid with 
\(400\times 400\times 400\) points, equalling \(6.4\cdot 10^7\)
evaluation points. Evaluation is again performed by applying
the fast multipole method and takes \(2.69\cdot10^4\)s using 32 
cores. The panel in the middle of Figure~\ref{fig:StanfordDragon} 
shows the reconstructed surface as zero level-set of the signed 
distance function, and computed by the ridge regression. The 
sparse reconstruction is depicted in the right panel. As can 
clearly be seen, the sparsity constraint amounts to a smoothing 
of the surface. Figure~\ref{fig:StanfordDragonCoeff} shows 
the clusters containing the active coefficients, which perfectly 
localize at the features of the implicit surface.
\subsection{Sparse reconstruction of temperature data}\label{ss:ERA} 
\begin{figure*}[t!]
\begin{center}
\begin{tikzpicture}
\draw(0,6)node{February};
\draw(5,6)node{July};
\draw(10,6)node{October};
\draw(-2.7,4.8)node[rotate=90]{relative error};
\draw(-2.7,2.4)node[rotate=90]{temperature};
\draw(-2.7,0)node[rotate=90]{contribution $k_1$};
\draw(-2.7,-2.4)node[rotate=90]{contribution $k_2$};
\draw(0,0)node{\includegraphics[scale=0.1,clip,trim=200 40 260 30]{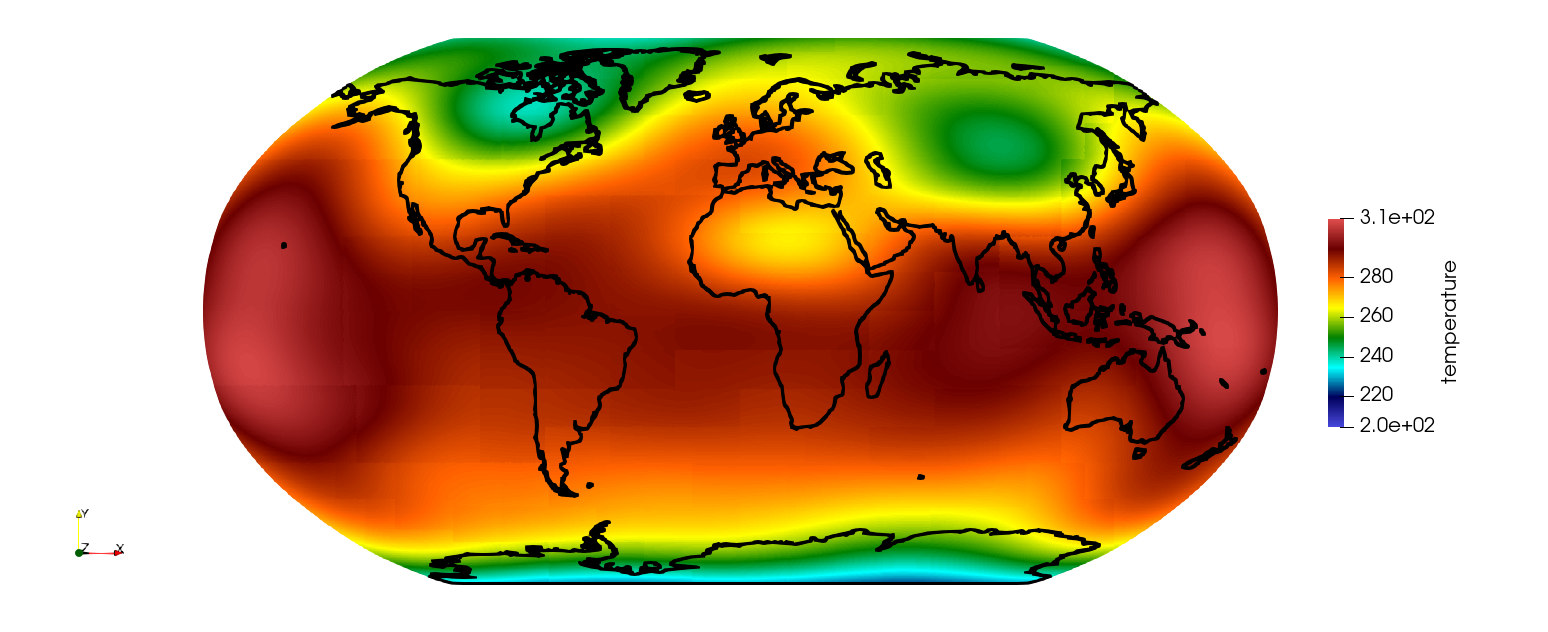}};
\draw(5,0)node{\includegraphics[scale=0.1,clip,trim=200 40 260 30]{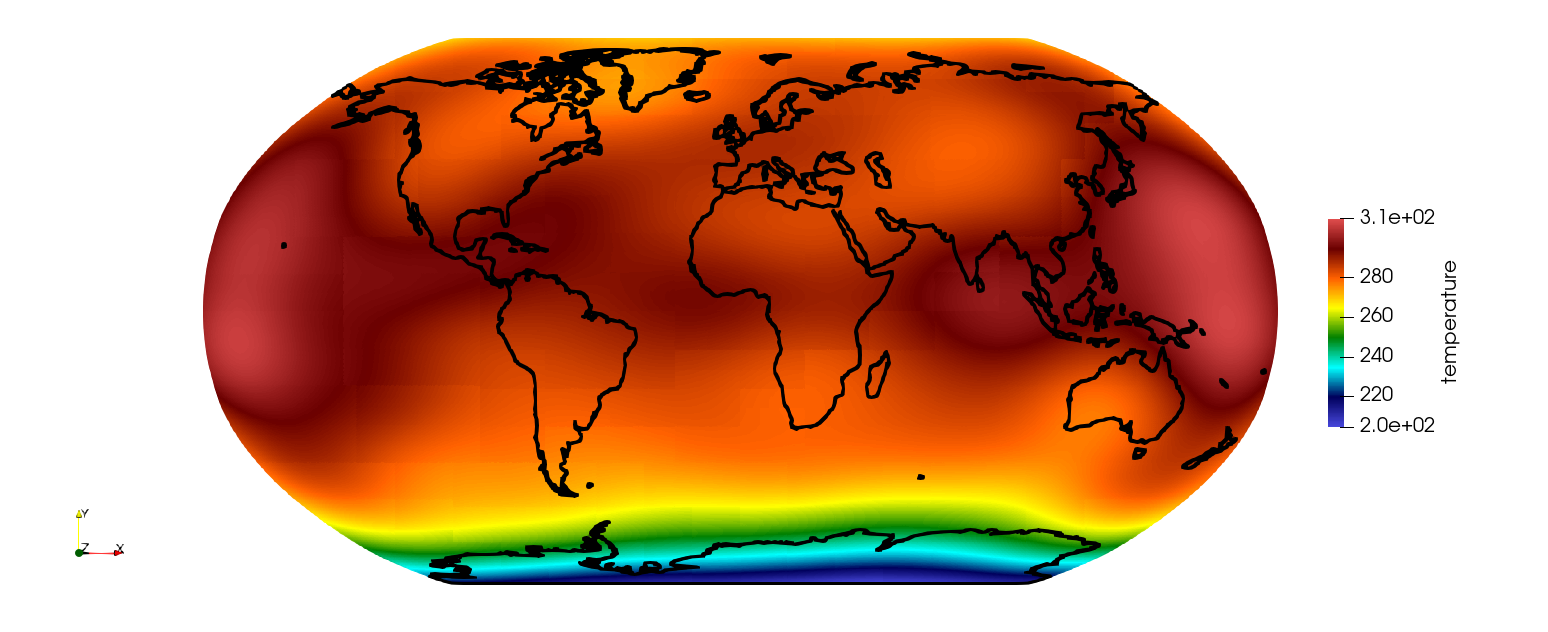}};
\draw(10,0)node{\includegraphics[scale=0.1,clip,trim=200 40 260 30]{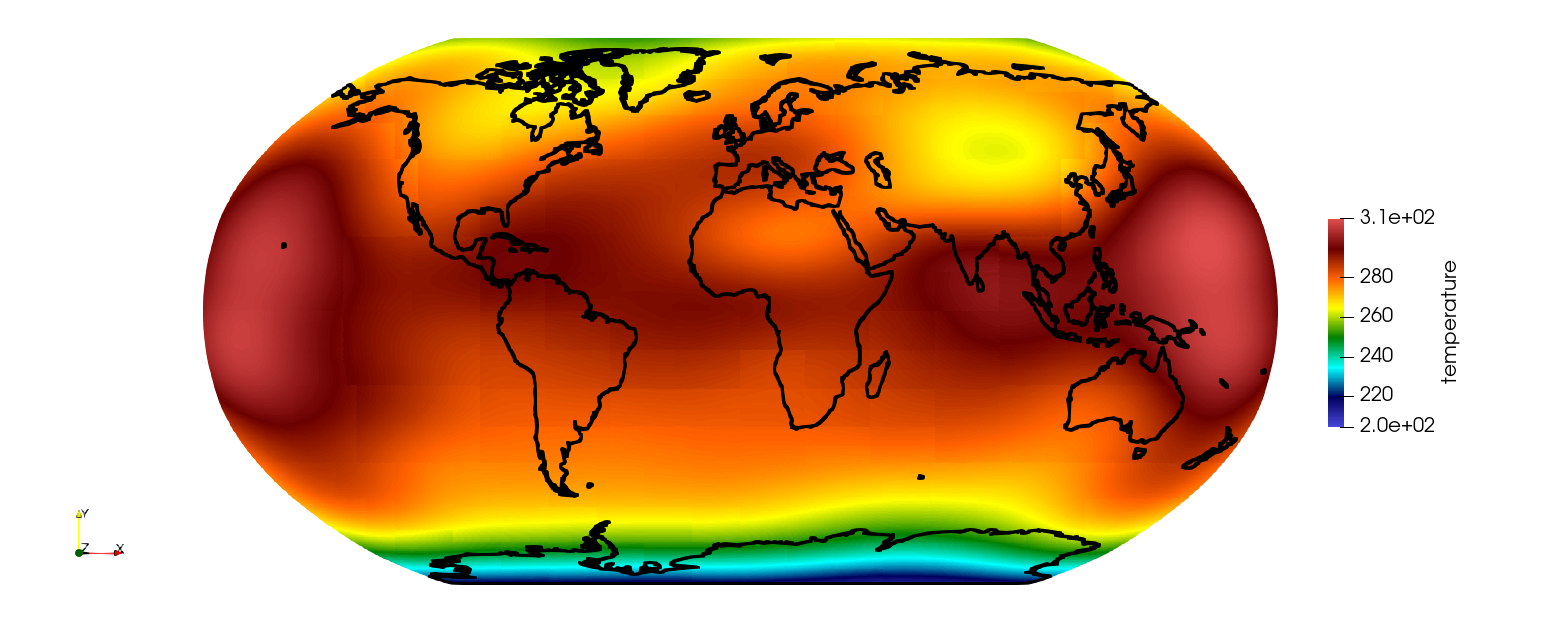}};
\draw(13.4,0)node{\includegraphics[scale=0.23,clip,trim=1320 190 70 195]{./Images/tempk1Oct.png}};
\draw(0,-2.4)node{\includegraphics[scale=0.1,clip,trim=200 40 260 30]{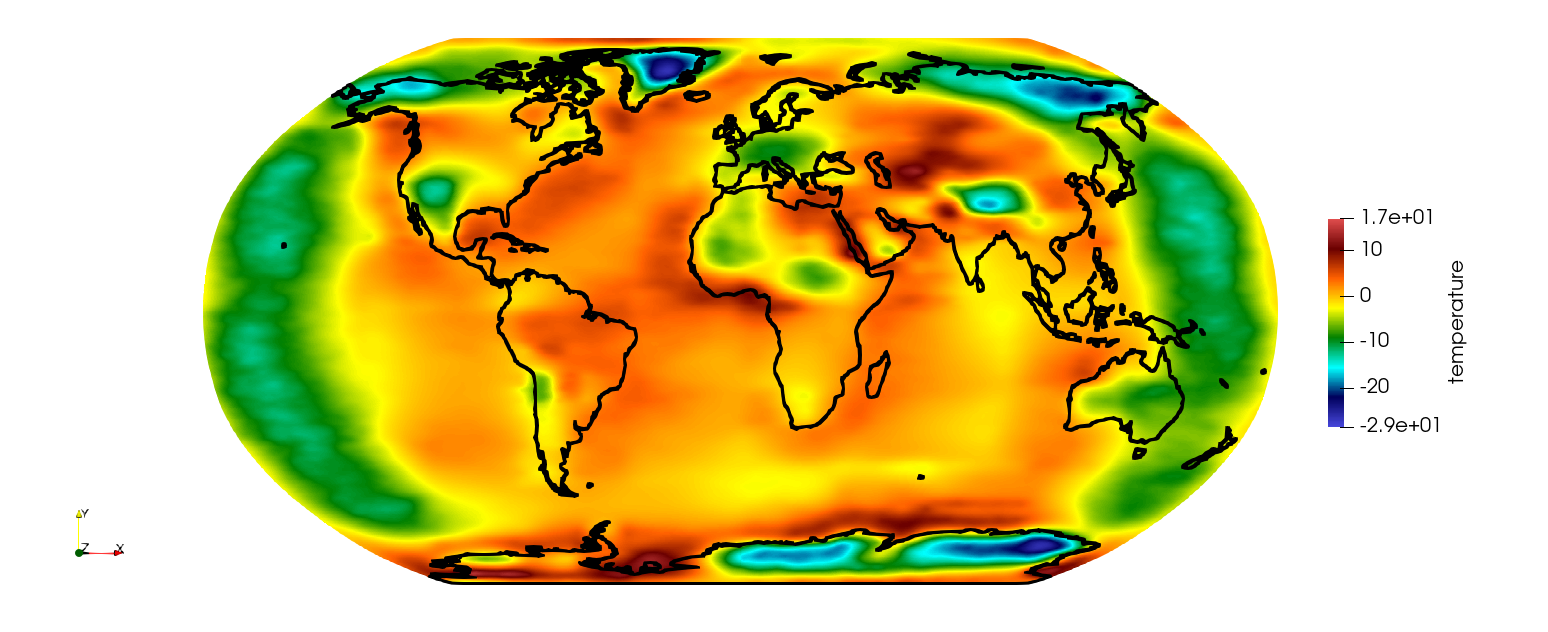}};
\draw(5,-2.4)node{\includegraphics[scale=0.1,clip,trim=200 40 260 30]{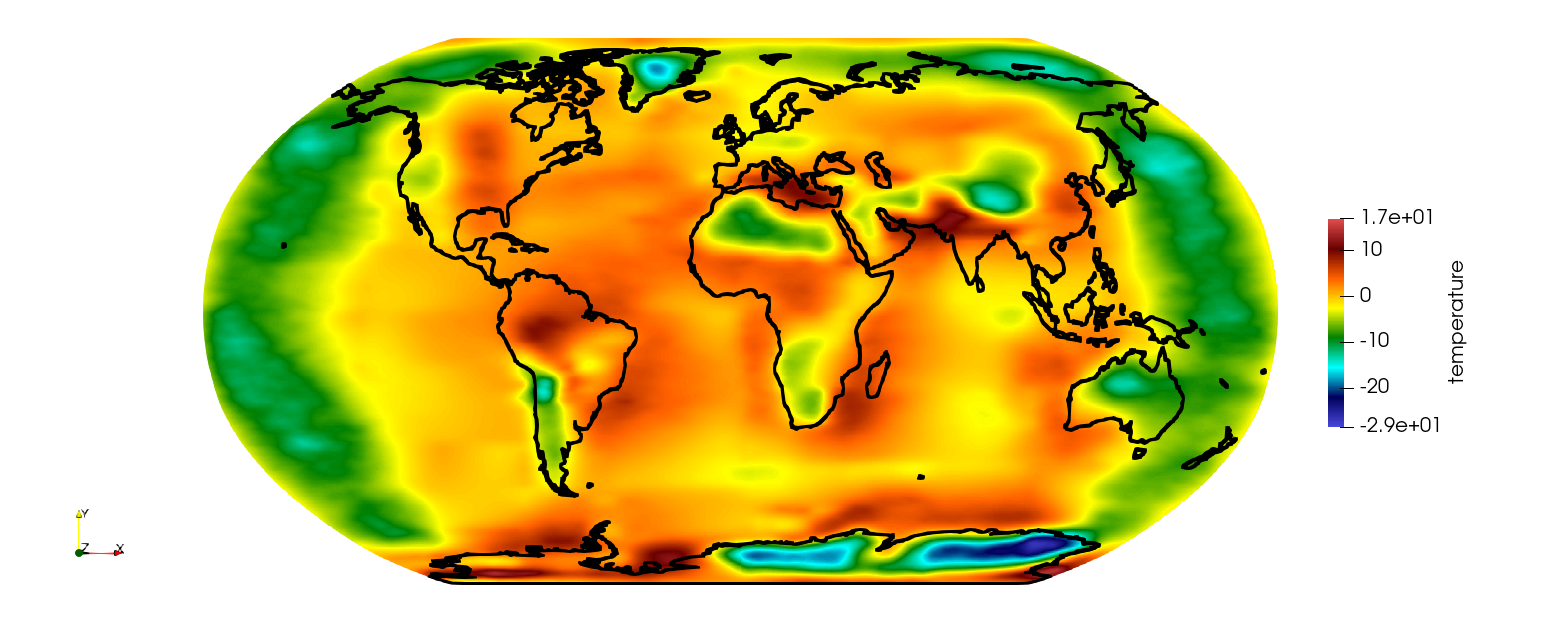}};
\draw(10,-2.4)node{\includegraphics[scale=0.1,clip,trim=200 40 260 30]{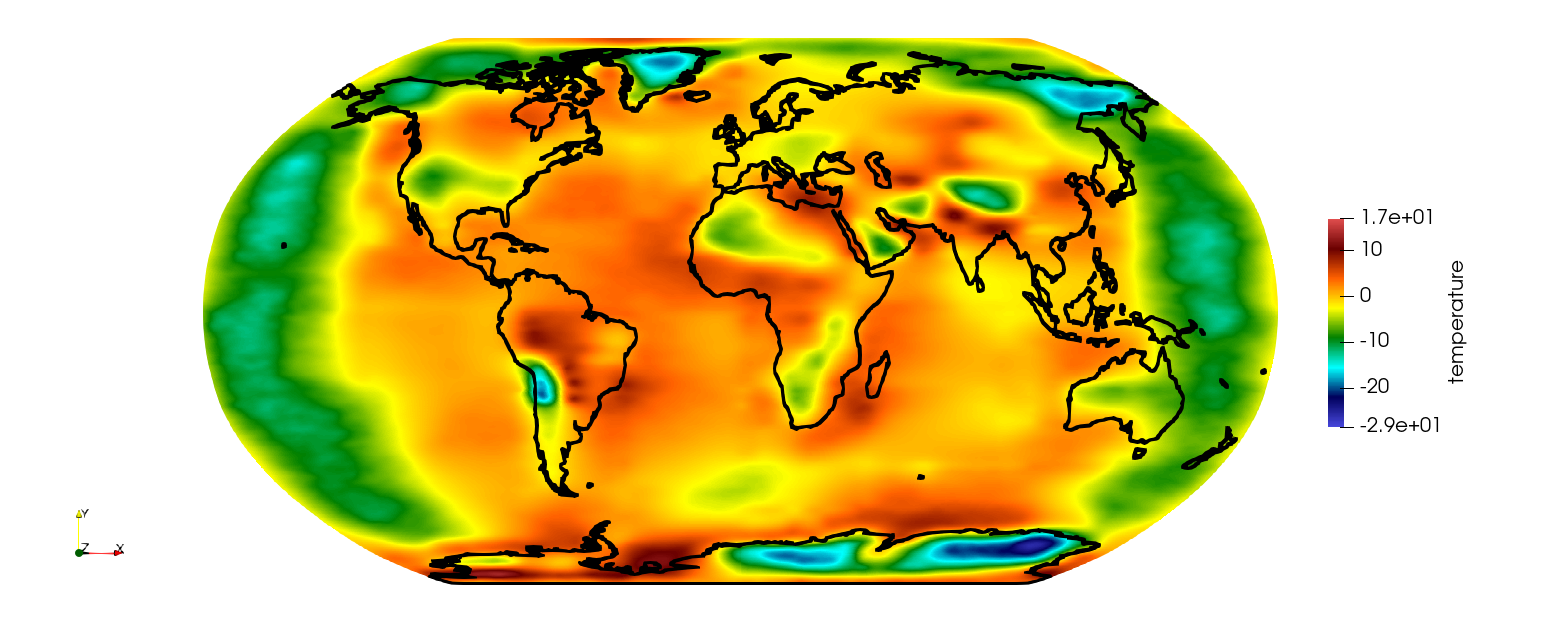}};
\draw(13.4,-2.4)node{\includegraphics[scale=0.23,clip,trim=1320 190 70 195]{./Images/tempk2Oct.png}};
\draw(0,2.4)node{\includegraphics[scale=0.1,clip,trim=200 40 260 30]{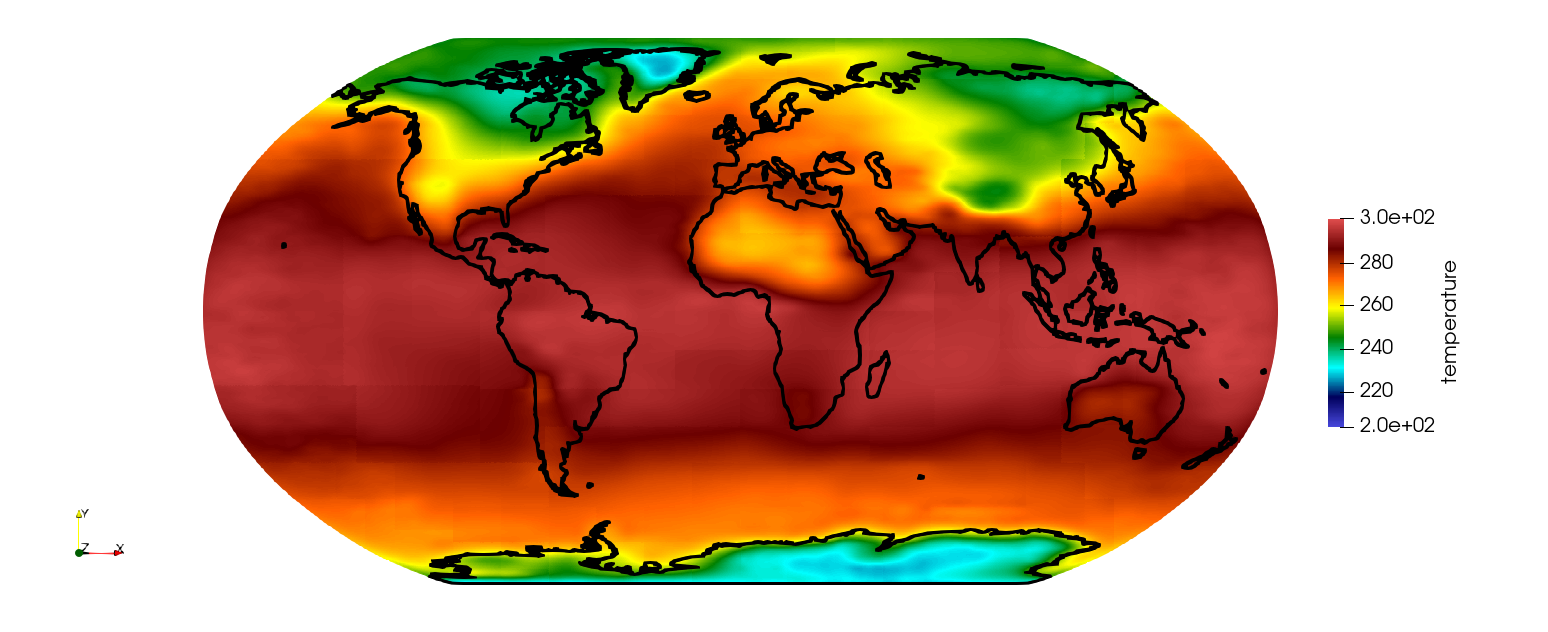}};
\draw(5,2.4)node{\includegraphics[scale=0.1,clip,trim=200 40 260 30]{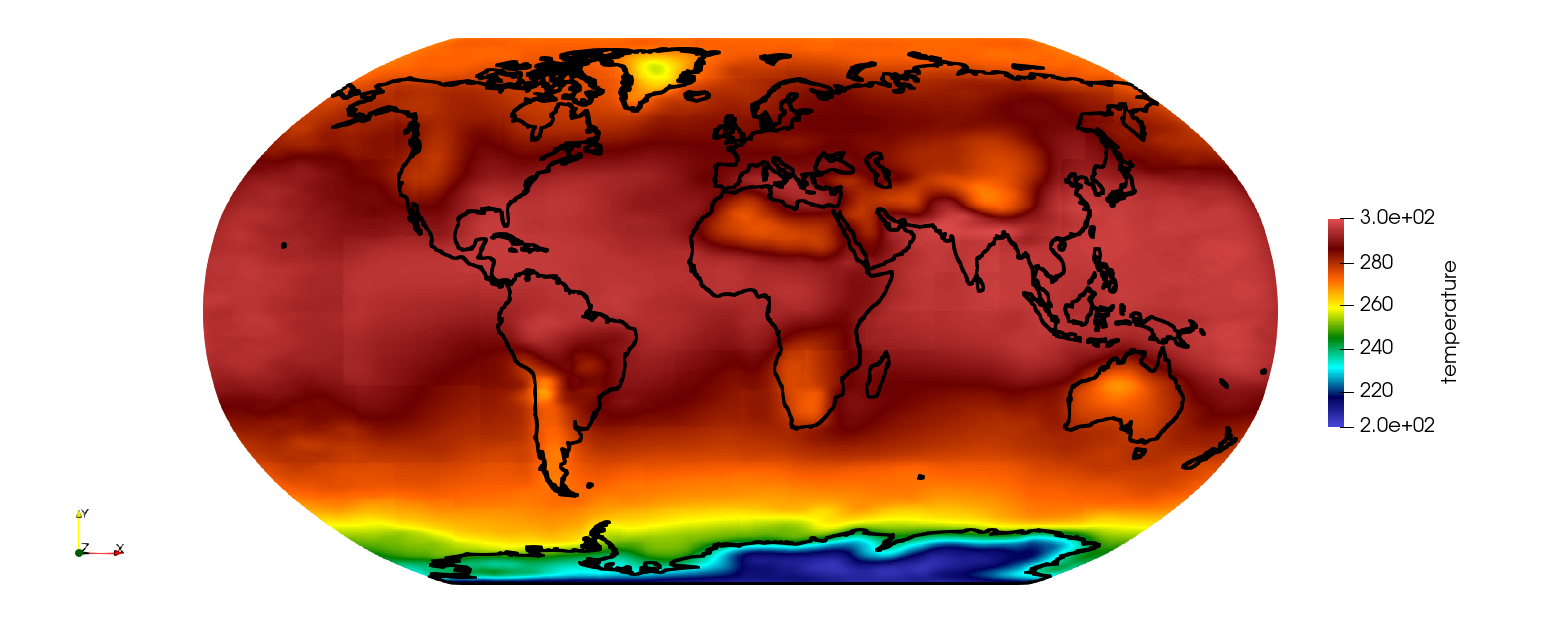}};
\draw(10,2.4)node{\includegraphics[scale=0.1,clip,trim=200 40 260 30]{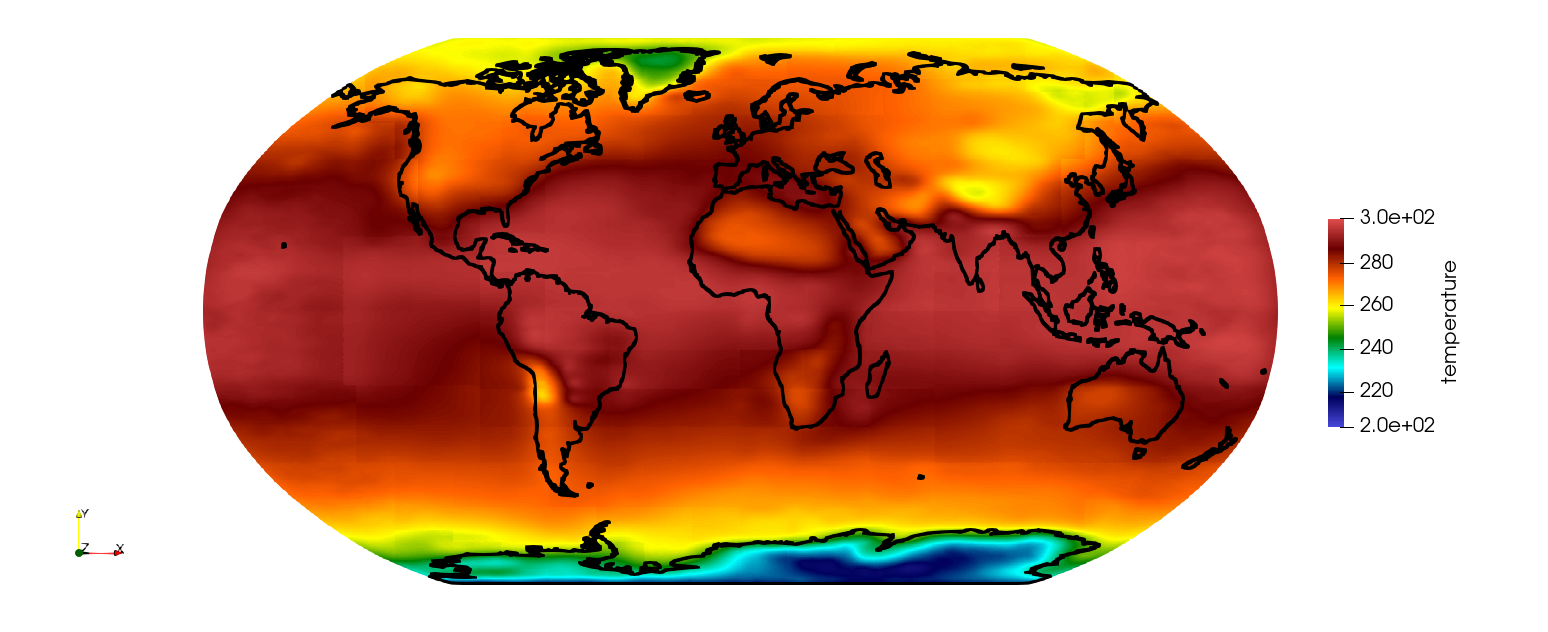}};
\draw(13.4,2.4)node{\includegraphics[scale=0.23,clip,trim=1320 190 70 195]{./Images/tempOct.png}};
\draw(0,4.8)node{\includegraphics[scale=0.1,clip,trim=200 40 260 30]{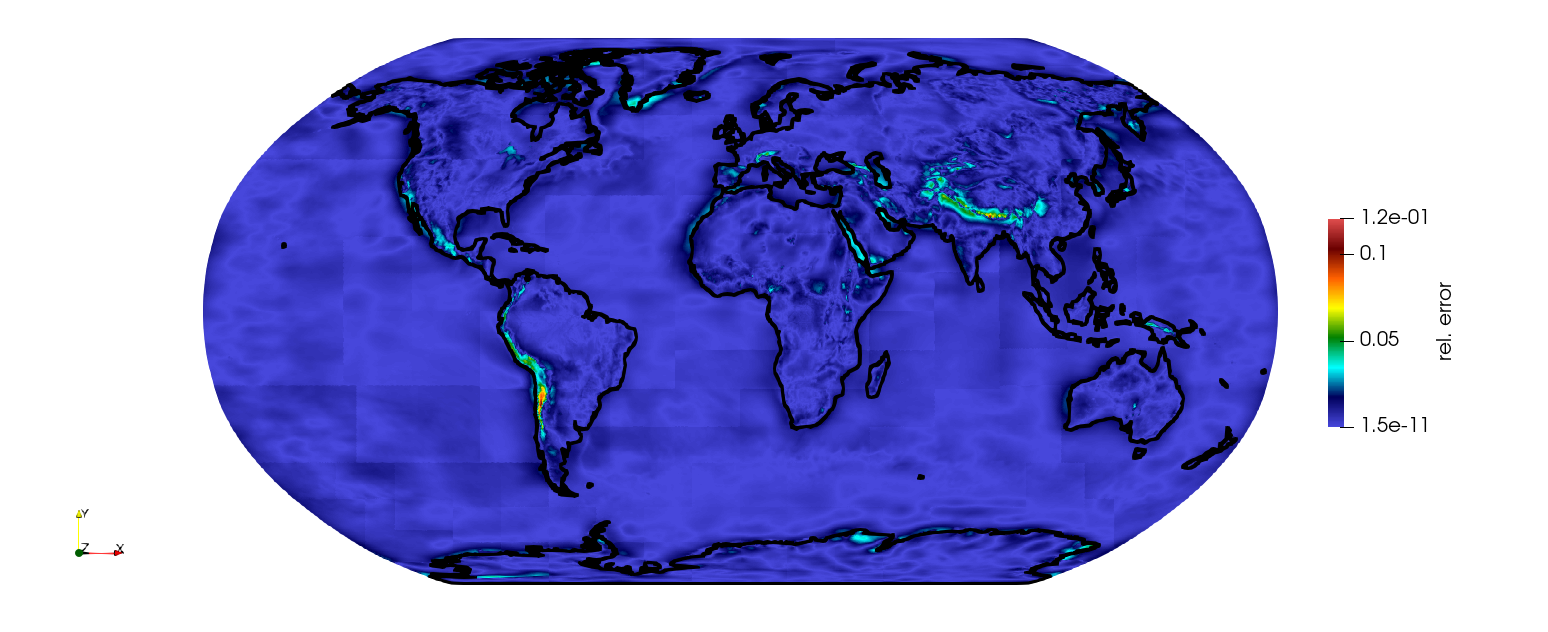}};
\draw(5,4.8)node{\includegraphics[scale=0.1,clip,trim=200 40 260 30]{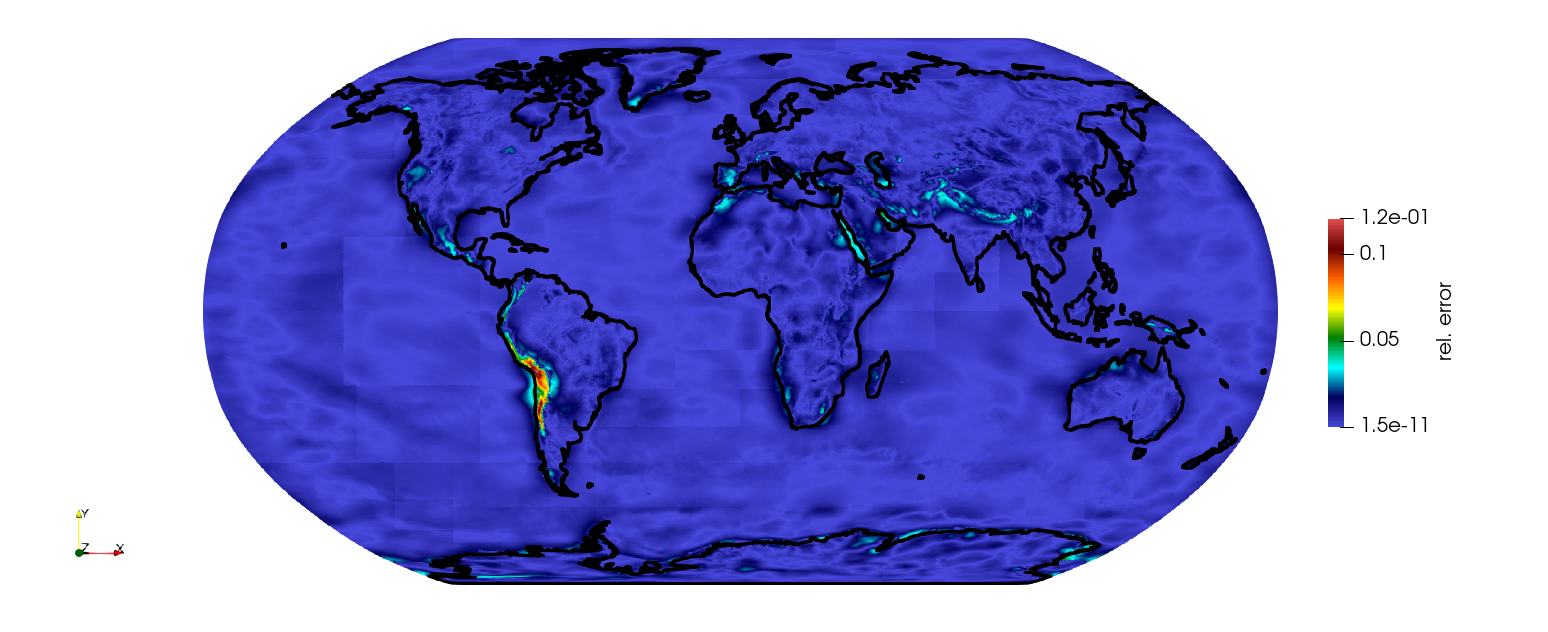}};
\draw(10,4.8)node{\includegraphics[scale=0.1,clip,trim=200 40 260 30]{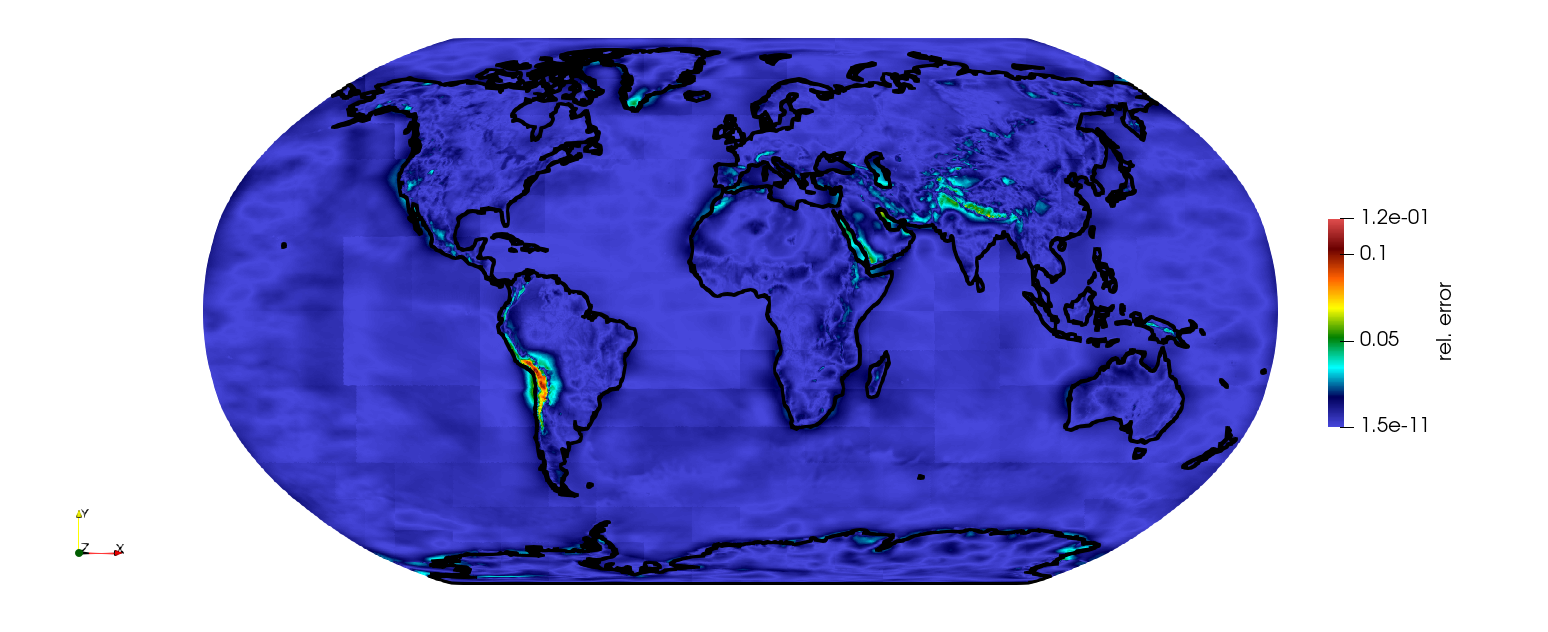}};
\draw(13.4,4.8)node{\includegraphics[scale=0.23,clip,trim=1320 190 70 195]{./Images/errOct.png}};
\end{tikzpicture}
\caption{\label{fig:Temperature}Sparse reconstruction of temperature data using two kernels.
The relative error is shown in the top row, the reconstructed temperature in the second row,
the contribution of \(k_1\) is seen in the third row and the contribution of \(k_2\) in the last row.}
\end{center}
\end{figure*}

In this numerical experiment, we perform the sparse 
reconstruction of temperature data. As dataset, we consider 
the monthly ERA5 temperature data set, which provides the 
temperature 2M above the surface. ERA5 is a reanalysis 
by the European Center for Medium-range Weather
Forecasts (ECMWF) of global climate and weather 
for the past eight decades. Data is available from 
1940 onwards, see \cite{ERA5}. We use the monthly 
data from 2022, which comprises 1\,038\,240 data points 
per month, resulting in 12\,458\,880 points in total. The 
original data format is World Geodesic System 1984 
(WGS 84), which we transform by using the Robinson 
projection, compare \cite{Rob74}, to reduce the distortion 
and the projection angles. From the data set we 
randomly select \(10\%\) of the points, 
resulting in \(N=1\,245\,888\) 
data points. The selection is performed
randomly in space-time. This means that a given point in
space might be present at a given time, but not at any other
time. Hence, the chosen locations are not persistent
in time and standard approaches like dynamic mode decomposition
(DMD), see, e.g., \cite{Sch10} or wavelets are not
directly applicable.

For the subsequent computation, the data 
are rescaled to the unit hyper-cube \([0,1]^3\).

We employ two kernels,
\[
\Kcal_1\big(({\bs x},t),({\bs y},t')\big)\isdef
k_{3/2}(\|{\bs x}-{\bs y}\|_2)k_{\text{per}}(|t-t'|)
\]
and
\[
\Kcal_2({\bs z},{\bs z}')\isdef k_{\exp}(\|{\bs z}-{\bs z}'\|_2),\
{\bs z}\isdef({\bs x},t),\ 
{\bs z}'\isdef({\bs y},t'),
\]
where we set the corresponding components according to
\begin{align*}
k_{3/2}(r)&\isdef(1+5\sqrt{3})e^{-5\sqrt{3}r},\\
k_{\text{per}}(r)&\isdef e^{-50\sin^2(\pi r)},\\
k_{\exp}(r)&\isdef e^{-100r}.
\end{align*}
The kernel \(\Kcal_1\) is a tensor product kernel
comprised of a Mat\'ern-3/2 kernel in space and a
periodic kernel in time, both with relatively large
correlation lengths. This kernel is intended to capture
the smooth parts of the temperature distribution over time.
The second kernel \(\Kcal_2\) is a very rough exponential 
kernel in space time and intended to capture sharp features.
The overall time for the compression of both kernels is 
$508$s. The compressed matrix contains for \(\Kcal_1\) 
contains $507$ entries per row on average and the 
estimated relative compression error is $6.02\cdot{10^{-5}}$ 
in the Frobenius norm. The compressed matrix contains 
for \(\Kcal_2\) contains $939$ entries per row on average 
and the estimated relative compression error is $3.37
\cdot{10^{-4}}$ in the Frobenius norm. We set \(w_i =
8\cdot 10^{-7}\) for \(i=1,\ldots,N\) and MRSSN is stopped 
if \(\|{\bs r}\|_\infty<9\cdot 10^{-7}\). The evaluation of the 
interpolant at all 12\,458\,880 points of the full data set
approximately takes between $50$s and $73$s per 
time step using 32 cores.

The top row of Figure~\ref{fig:Temperature} shows the 
reconstruction error evaluated at the full data set for the 
months February, July, and October. The relative error 
is everywhere smaller than 12\% and strongly localized 
at the mountains (Andes and Himalayas). The second 
row shows the reconstructed temperature distribution 
evaluated at the full data set. The third row shows the 
contribution of \(\Kcal_1\), where \(\|{\bs\alpha}_1\|_0=712\).
The kernel \(\Kcal_1\) indeed perfectly captures the 
coarse-scale structure of the temperature distribution.
The last row shows the contribution of \(\Kcal_2\), 
where \(\|{\bs\alpha}_2\|_0=5430\). This kernel localizes 
coast lines and mountains and thus actually captures the
sharp features. Moreover, it accounts for the periodicity 
in longitudinal direction. 

We finally remark that it would
also be possible to directly work with the temperature 
data mapped to the earth's surface, which would 
however come at the cost of a four dimensional 
approach instead of a three dimensional one.

\section{Conclusion}\label{sct:conclusion}
In this article, we have presented the novel concept of samplet 
basis pursuit. Samplets constitute a multiresolution basis tailored 
to scattered data. This renders them the method of choice for
feature detection and sparse reconstruction of such data. 
Interpolation is directly facilitated by embedding of samplets 
into the reproducing kernel Hilbert space context. An efficient 
computational algorithm for the computation of sparse 
representations is obtained by employing the multiresolution 
semi-smooth Newton method in samplet coordinates. Besides 
a benchmark problem, we have presented extensive numerical 
studies for the reconstruction of implicit surfaces from 
measurements of the signed distance function and for the 
reconstruction of temperature data from scattered measurements.
In particular, samplets naturally allow for the approximation 
of scattered data in 
space-time and are therefore also able to capture dynamic 
features.

\section*{Acknowledgements}
DB was funded by the Swiss Federal Office of Energy SFOE as part 
of the SWEET project SURE. HH was funded in parts by the SNSF 
by the grant ``Adaptive Boundary Element Methods Using Anisotropic
Wavelets'' (200021\_192041). MM was funded in parts by the SNSF 
starting grant ``Multiresolution methods for unstructured data'' (TMSGI2\_211684).

\bibliographystyle{plain}
\bibliography{literature}

\begin{thebibliography}{10}

\bibitem{FISTA}
A.~Beck and M.~Teboulle.
\newblock A fast iterative shrinkage-thresholding algorithm for linear inverse
  problems.
\newblock {\em SIAM J. Imaging Sci.}, 2(1):183--202, 2009.

\bibitem{BNS97}
B.~Blaschke, A.~Neubauer, and O.~Scherzer.
\newblock On convergence rates for the iteratively regularized gauss-newton
  method.
\newblock {\em IMA J. Numer. Anal.}, 17(3):421--436, 1997.

\bibitem{ADMM}
S.~Boyd, N.~Parikh, E.~Chu, B.~Peleato, and J.~Eckstein.
\newblock Distributed optimization and statistical learning via the alternating
  direction method of multipliers.
\newblock {\em Found. Trends Mach. Learn.}, 3(1):1–122, 2010.

\bibitem{Candes}
E.~Cand{\`e}s, J.~Romberg, and T.~Tao.
\newblock Stable signal recovery from incomplete and inaccurate measurements.
\newblock {\em Comm. Pure Appl. Math.}, 59(8):1207--1223, 2006.

\bibitem{curvelets}
E.J. Candes and D.L. Donoho.
\newblock Curvelets: {A} surprisingly effective nonadaptive representation for
  objects with edges.
\newblock In A.~Cohen, C.~Rabut, and L.~Schumaker, editors, {\em Curves and
  Surface Fitting: Saint-Malo 1999}, page 105–120, Nashville, 2000.
  Vanderbilt University Press.

\bibitem{CarrEtAl01}
J.C. Carr, R.K. Beatson, J.B. Cherrie, T.J. Mitchell, W.R. Fright, B.C.
  McCallum, and T.R. Evans.
\newblock Reconstruction and representation of {3D} objects with radial basis
  functions.
\newblock In {\em Proceedings of the 28th annual conference on Computer
  graphics and interactive techniques}, SIGGRAPH '01, pages 67--76, New York,
  2001. Association for Computing Machinery.

\bibitem{BasisPursuit}
S.~Chen and D.L. Donoho.
\newblock Basis pursuit.
\newblock In {\em Proceedings of 1994 28th Asilomar Conference on Signals,
  Systems and Computers}, volume~1, pages 41--44, Pacific Grove, CA, USA, 1994.
  IEEE.

\bibitem{CDS98}
S.S. Chen, D.L. Donoho, and M.A. Saunders.
\newblock Atomic decomposition by basis pursuit.
\newblock {\em SIAM J. Sci. Comput.}, 20(1):33--61, 1998.

\bibitem{CZQ}
X.~Chen, Z.~Nashed, and L.~Qi.
\newblock Smoothing methods and semismooth methods for nondifferentiable
  operator equations.
\newblock {\em SIAM J. Numer. Anal.}, 38(4):1200–1216, 2000.

\bibitem{ISTA}
I.~Daubechies, M.~Defrise, and C.~De~Mol.
\newblock An iterative thresholding algorithm for linear inverse problems with
  a sparsity constraint.
\newblock {\em {C}omm. {P}ure {A}ppl. {M}ath.}, 57:1413--1457, 2004.

\bibitem{contourlets}
M.N. Do and M.~Vetterli.
\newblock The contourlet transform: an efficient directional multiresolution
  image representation.
\newblock {\em IEEE Trans. Image Proc.}, 14:2091--2106, 2005.

\bibitem{Donoho}
D.L. Donoho.
\newblock Compressed sensing.
\newblock {\em IEEE Trans. Inf. Theory}, 52(4):1289--1306, 2006.

\bibitem{Fasshauer2007}
G.E. Fasshauer.
\newblock {\em Meshfree approximation methods with {MATLAB}}.
\newblock World Scientific, River Edge, 2007.

\bibitem{rauhut}
S.~Foucart and H.~Rauhut.
\newblock {\em A Mathematical Introduction to Compressive Sensing}.
\newblock Applied and Numerical Harmonic Analysis. Birkh\"auser, New York,
  2013.

\bibitem{LorenzGriesse}
R.~Griesse and D.A. Lorenz.
\newblock A semismooth {N}ewton method for {T}ikhonov functionals with sparsity
  constraints.
\newblock {\em Inverse Problems}, 24(3):035007, 2008.

\bibitem{GCG05}
V.~Guigue, A.~Rakotomamonjy, and S.~Canu.
\newblock Kernel basis pursuit.
\newblock In J.~Gama, R.~Camacho, P.~B. Brazdil, A.~M. Jorge, and L.~Torgo,
  editors, {\em Machine Learning: ECML 2005}, pages 146--157, Berlin,
  Heidelberg, 2005. Springer.

\bibitem{shearlets}
K.~Guo and D.~Labate.
\newblock Optimally sparse multidimensional representation using shearlets.
\newblock {\em SIAM J. Math. Anal.}, 39(1):298--318, 2007.

\bibitem{HKS05}
H.~Harbrecht, U.~K{\"a}hler, and R.~Schneider.
\newblock Wavelet {G}alerkin {BEM} on unstructured meshes.
\newblock {\em Comput. Vis. Sci.}, 8(3--4):189--199, 2005.

\bibitem{HM22}
H.~Harbrecht and M.~Multerer.
\newblock Samplets: Construction and scattered data compression.
\newblock {\em J.\ Comput.\ Phys.}, 471:111616, 2022.

\bibitem{HMSS22}
H.~Harbrecht, M.~Multerer, O.~Schenk, and C.~Schwab.
\newblock Multiresolution kernel matrix algebra.
\newblock {\em arXiv-Preprint}, arXiv:2211.11681, 2022.
\newblock to appear in Numer. Math.

\bibitem{ERA5}
H.~Hersbach, B.~Bell, P.~Berrisford, S.~Hirahara, A.~Hor{\'a}nyi,
  J.~Muñoz-Sabater, J.~Nicolas, C.~Peubey, R.~Radu, D.~Schepers, A.~Simmons,
  C.~Soci, S.~Abdalla, X.~Abellan, G.~Balsamo, P.~Bechtold, G.~Biavati,
  J.~Bidlot, M.~Bonavita, G.~De~Chiara, P.~Dahlgren, D.~Dee, M.~Diamantakis,
  R.~Dragani, J.~Flemming, R.~Forbes, M.~Fuentes, A.~Geer, L.~Haimberger,
  S.~Healy, R.~J. Hogan, E.~Hólm, M.~Janisková, S.~Keeley, P.~Laloyaux,
  P.~Lopez, C.~Lupu, G.~Radnoti, P.~de~Rosnay, I.~Rozum, F.~Vamborg,
  S.~Villaume, and J.-N. Thépaut.
\newblock The era5 global reanalysis.
\newblock {\em Q. J. R. Meteorol.}, 146(730):1999--2049, 2020.

\bibitem{HIK}
M.~Hinterm\"uller, K.~Ito, and K.~Kunisch.
\newblock The primal-dual active set strategy as a semismooth {N}ewton method.
\newblock {\em SIAM J.~Optim.}, 13:865--888, 2003.

\bibitem{learning}
G.~James, D.~Witten, T.~Hastie, and R.~Tibshirani.
\newblock {\em An {I}ntroduction to {S}tatistical {L}earning}.
\newblock Springer Texts in Statistics. Springer, New York, 2013.

\bibitem{Koenig}
H.~K\"onig.
\newblock {\em Eigenvalue distribution of compact operators}, volume~16 of {\em
  Operator Theory: Advances and Applications}.
\newblock Birkhäuser, Basel, 1986.

\bibitem{Lorenz}
D.A. Lorenz.
\newblock Convergence rates and source conditions for tikhonov regularization
  with sparsity constraints.
\newblock {\em J. Inverse Ill-Posed Probl.}, 16(5):463--478, 2008.

\bibitem{MH}
S.G. Mallat and W.L. Hwang.
\newblock Singularity detection and processing with wavelets.
\newblock {\em IEEE Trans. Inf. Theory}, 38(2):617--643, 1992.

\bibitem{MZ}
S.G. Mallat and Z.~Zhang.
\newblock Matching pursuits with time-frequency dictionaries.
\newblock {\em IEEE Trans. Sign. Proc.}, 41(12):3397--3415, 1993.

\bibitem{RT}
R.~Ramlau and G.~Teschke.
\newblock A {T}ikhonov-based projection iteration for nonlinear ill-posed
  problems with sparsity constraints.
\newblock {\em Numer. Math.}, 104(2):177–203, 2006.

\bibitem{Rob74}
A.H. Robinson.
\newblock A new map projection: Its development and characteristics.
\newblock {\em International yearbook of cartography}, 14(1974):145--155, 1974.

\bibitem{Sch10}
P.~J. Schmid.
\newblock Dynamic mode decomposition of numerical and experimental data.
\newblock {\em J. Fluid Mech.}, 656:5--28, 2010.

\bibitem{Tao}
T.~Tao and E.J. Cand{\`e}s.
\newblock Near-optimal signal recovery from random projections: universal
  encoding strategies?
\newblock {\em IEEE Trans. Inf. Theory}, 52(12):5406–5425, 2006.

\bibitem{TW03}
J.~Tausch and J.~White.
\newblock Multiscale bases for the sparse representation of boundary integral
  operators on complex geometry.
\newblock {\em SIAM J. Sci. Comput.}, 24(5):1610--1629, 2003.

\bibitem{tropp}
J.A. Tropp.
\newblock Greed is good: algorithmic results for sparse approximation.
\newblock {\em IEEE Trans. Inf. Theory}, 50(10):2231--2242, 2004.

\bibitem{Wendland2004}
H.~Wendland.
\newblock {\em Scattered Data Approximation}.
\newblock Cambridge University Press, Cambridge, 2004.

\end{thebibliography}
\end{document}